\documentclass{article}

\usepackage[main, final]{neurips_2026}

\usepackage[utf8]{inputenc} % allow utf-8 input
\usepackage[T1]{fontenc}    % use 8-bit T1 fonts
\usepackage{hyperref}       % hyperlinks
\usepackage{url}            % simple URL typesetting
\usepackage{booktabs}       % professional-quality tables
\usepackage{amsfonts}       % blackboard math symbols
\usepackage{nicefrac}       % compact symbols for 1/2, etc.
\usepackage{microtype}      % microtypography
\usepackage{xcolor}         % colors
\usepackage{amsmath}
\usepackage[capitalise,noabbrev,nameinlink]{cleveref}
\usepackage{graphicx}
\usepackage{multirow}
\usepackage{setspace}
\usepackage{caption}
\usepackage{float}

\newcommand{\eg}{e.g.}

\newcommand{\Zadj}{\raisebox{-0.1ex}{\ensuremath{\mathbf{Z}}}}
\newcommand{\Zpadj}{\raisebox{-0.1ex}{\ensuremath{\mathbf{Z}'}}}
\newcommand{\sm}[1]{{\tiny #1}}

\title{Improving Flexible Image Tokenizers for Autoregressive Image Generation}

\author{%
    Zixuan Fu$^{1}$ \quad Lanqing Guo$^{2}$ \quad Chong Wang$^{1}$ \quad
    Binbin Song$^{3}$ \quad Ding Liu$^{4}$ \quad Bihan Wen$^{1}$\thanks{Corresponding author.} \\
    $^{1}$Nanyang Technological University \quad
    $^{2}$The University of Texas at Austin \\
    $^{3}$Harbin Institute of Technology \quad
    $^{4}$Meta AI \\
    \texttt{\{zixuan006,wang1711,bihan.wen\}@ntu.edu.sg},
    \texttt{lanqing.guo@austin.utexas.edu} \\
    \texttt{binbinsong@hit.edu.cn},
    \texttt{Liudingdavy@gmail.com}
}

\begin{document}

\maketitle

\begin{abstract}
Flexible image tokenizers aim to represent an image using an ordered 1D variable-length token sequence. 
This flexible tokenization is typically achieved through nested dropout, where a portion of trailing tokens is randomly truncated during training, and the image is reconstructed using the remaining preceding sequence.
However, this tail-truncation strategy inherently concentrates the image information in the early tokens, limiting the effectiveness of downstream AutoRegressive (AR) image generation as the token length increases.
To overcome these limitations, we propose \textbf{ReTok}, a flexible tokenizer with \underline{Re}dundant \underline{Tok}en Padding and Hierarchical Semantic Regularization, designed to fully exploit all tokens for enhanced latent modeling.
Specifically, we introduce \textbf{Redundant Token Padding} to activate tail tokens more frequently, thereby alleviating information over-concentration in early tokens. 
In addition, we apply \textbf{Hierarchical Semantic Regularization} to align the decoding features of earlier tokens with those from a pre-trained vision foundation model, while progressively reducing the regularization strength toward the tail to allow finer low-level detail reconstruction. 
Extensive experiments demonstrate the effectiveness of ReTok: on ImageNet 256$\times$256, our method achieves superior generation performance compared with both flexible and fixed-length tokenizers. Code is available at: \href{https://github.com/zfu006/ReTok}{https://github.com/zfu006/ReTok}
\end{abstract}

\section{Introduction}

\label{sec:intro}
Autoregressive (AR) models have demonstrated remarkable capability in image generation~\cite{esser2021taming,sun2024autoregressive,tian2024visual,wang2025simplear,yu2023language}, owing to their inherent flexibility, scalability \cite{grattafiori2024llama, chowdhery2023palm, brown2020language, radford2018improving}, and potential for extension into unified multimodal frameworks~\cite{chen2025janus, team2024chameleon, wu2024vila, wang2024emu3}.
%
%Generally, AR image generation models rely on a visual tokenizer to compress images from the pixel space into a compact, discrete latent space, where they then model the image distribution in the latent space through next-token prediction.
Typically, AR image generators rely on a visual tokenizer to compress images from the pixel space into a compact, discrete latent space, where the image distribution is modeled through next-token prediction.
%
%Therefore, the visual tokenizer is crucial for downstream AR modeling, and the generation quality of AR models has been significantly boosted by recent progress in visual tokenizers \cite{xiong2025gigatok, yu2023language, van2017neural, esser2021taming, chang2022maskgit, weber2024maskbit, yu2024image}.
As a result, the visual tokenizer is crucial for downstream AR modeling, and advances in tokenizer design have greatly boosted generation quality~\cite{xiong2025gigatok, yu2023language, van2017neural, esser2021taming, chang2022maskgit, weber2024maskbit, yu2024image}.

Despite this progress, most visual tokenizers encode images into fixed-grid 1D or 2D representations, which fail to capture the heterogeneous complexity of natural images and constrain the flexibility of AR models.
%
% From a compression perspective, an image should be dynamically compressed into a varying number of tokens based on the complexity of its objects and scene. 
% %
% Meanwhile, from the AR generation perspective, AR models should be capable of generating images with variable-length sequences, enabling a more flexible and resource-efficient synthesis process.
%
To overcome these limitations, \textbf{flexible tokenizers} are proposed to represent an image as a 1D variable-length token sequence~\cite{bachmann2025flextok, miwa2025one, duggal2024adaptive, yan2024elastictok, huang2025spectralar, liu2025detailflow, patel2026vibetoken, wen2025principal}.
%
%During training, nested dropout \cite{rippel2014learning} is applied to truncate the tail tokens while tokenizers are encouraged to reconstruct the image with the remaining early tokens.  
During training, these tokenizers adopt nested dropout~\cite{rippel2014learning}, where trailing tokens are randomly truncated, and the image is reconstructed using the remaining prefix tokens.
%
%After that, AR models trained under the flexible tokenizers are capable of generating tokens with variable lengths, which are then decoded into plausible images through the tokenizer decoder.
This enables AR models to generate variable-length token sequences that can be decoded into plausible images via the tokenizer’s decoder.
%
%While their effectiveness, we find that the AR models, building upon the flexible tokenizers, fail to consistently improve their generation quality as the generated sequence length grows.
However, despite their flexibility, we observe that AR models built upon such tokenizers fail to \textbf{consistently} improve generation quality as the sequence length increases.
%
%Moreover, the upper bound generation quality of those AR generators significantly lags behind their counterparts that are trained on the fixed-length tokenizer under the same token budget.
Moreover, their generation performance remains significantly lower than that of AR models trained with fixed-length tokenizers under the same token budget.
%
%We identify the generation bottleneck of flexible tokenizers in \cref{fig_bottleneck}: generation quality (gFID) achieves negligible gains when increasing tokens (\eg, from 128 tokens to 256 tokens).
As illustrated in \cref{fig2_bottleneck}, the generation quality (measured by gFID) shows negligible improvement even when increasing the token count (\eg, from 128 to 256 tokens), revealing a fundamental bottleneck in current flexible tokenization strategies.
\begin{figure}[t!]
    \centering
    \includegraphics[width=\linewidth]{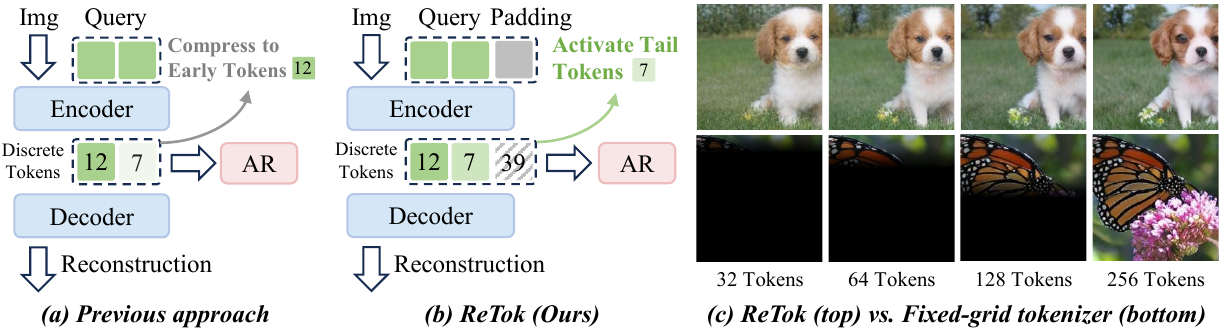}
    \caption{Overview of our method. (a) Previous flexible tokenizers (\eg, FlexTok \cite{bachmann2025flextok} and One-D-Piece \cite{miwa2025one}) with naive nested dropout compress information into early tokens, compromising generation quality. (b) Our method with redundant token padding activates tail tokens, consistently improving generation quality as the sequence extends. (c) Comparison of ReTok with fixed-length tokenizers \cite{sun2024autoregressive} in Autoregressive (AR) generation. Unlike fixed-length tokenizers that require generating a fixed-length sequence before decoding, ReTok allows AR models to produce variable-length sequences.}
    \label{fig_overview}
\end{figure}
In this paper, we propose \textbf{ReTok}, a 1D flexible tokenizer equipped with two novel training strategies that effectively address the limitations of existing flexible tokenizers.  
%
% We first improve the naive nested dropout by padding redundant tokens at the sequence tail, encouraging more activations of the trailing tokens at the original sequence. 
%
%We first introduce redundant token padding by concatenating additional tokens to the sequence tail, facilitating the utilization of the trailing tokens at the original sequence when applying nested dropout.
Specifically, we introduce \textbf{redundant token padding}, which appends additional tokens to the sequence tail to increase the activation frequency of trailing tokens during nested dropout, thereby promoting more balanced information distribution across the sequence.
%
%We further employ \textbf{hierarchical semantic regularization} to align the tokenizer's decoding features of earlier tokens with the semantic features extracted by a pre-trained vision foundation model (\eg, DINOv2 \cite{oquab2023dinov2}), and progressively decay the constraint for the subsequent tokens to reconstruct fine details of images.
We further employ \textbf{hierarchical semantic regularization}, which aligns the decoding features of earlier tokens with high-level semantic representations from a pre-trained vision foundation model, \eg, DINOv2~\cite{oquab2023dinov2}, while progressively decaying the constraint toward the later tokens to enable the reconstruction of fine-grained visual details.
%
%To this end, our proposed ReTok greatly alleviates the bottleneck of flexible tokenizers for the downstream AR generation, thereby advancing the practical application of the flexible tokenizers.
Together, these designs substantially alleviate the generation bottleneck of flexible tokenizers and enhance the downstream AR image generation performance, advancing the practical adoption of flexible tokenization.

Our main contributions are summarized as follows:
\begin{itemize}
    \item 
    %We propose \textbf{ReTok}, a novel 1D visual tokenizer that allows flexible AR generation, significantly improving the generation bottleneck of flexible tokenizers.
    We propose \textbf{ReTok}, a novel 1D visual tokenizer that significantly mitigates the generation bottleneck of existing flexible tokenizers and enables high-quality AR generation.
    \item
    We introduce \textbf{redundant token padding} and \textbf{hierarchical semantic regularization}, which allow the tokenizer to fully exploit every token in the sequence and achieve consistent improvements in generation quality with longer token lengths.
    \item
    %We conduct extensive experiments to verify the effectiveness of ReTok. Our method demonstrates superior generation performance among flexible tokenizers, while achieving comparable quality to that of fixed-length 1D tokenizers.
    We conduct extensive experiments demonstrating that ReTok achieves superior generation performance among flexible tokenizers and attains comparable quality to state-of-the-art fixed-length 1D tokenizers.
\end{itemize}

\section{Related Work}
\label{sec:related_work}

\subsection{Image Tokenizer}
\noindent \textbf{Fixed-grid 1D and 2D Image Tokenizers.} 
The objective of image tokenizers is to compress images into a compact latent space, which can then be modeled using generative models. 
For autoregressive image modeling, discrete tokenizers are widely adopted. 
VQVAE \cite{van2017neural} first introduces vector quantization for discrete image modeling, while VQGAN \cite{esser2021taming} further improves the perceptual quality of reconstructed images by incorporating the adversarial and perceptual losses \cite{goodfellow2020generative, johnson2016perceptual}.
Recent works have further advanced the development of discrete tokenizers through various improvements, such as replacing convolutional architectures with Vision Transformers (ViTs) \cite{yu2021vector, cao2023efficient}, scaling up the codebook size \cite{zhu2024scaling, qu2025tokenflow, ma2025unitok}, refining quantization strategies \cite{mentzer2023finite, zhu2024scaling, zhao2024image, yu2023language, weber2024maskbit}, and introducing multi-scale residual quantization \cite{tian2024visual, han2025infinity}, among others. 

While previous tokenizers encode images into a 2D grid with spatial structures, recent 1D tokenizers further eliminate this inductive bias by encoding images into a 1D sequence \cite{chen2025masked, chen2025softvq, yu2024image, xiong2025gigatok, zheng2025vision}. 
TiTok \cite{yu2024image}, a ViT-based 1D tokenizer, initializes query tokens at the encoder.
These query tokens and image patch embeddings are then jointly fed into the ViT encoder to learn the image latent representation.
After that, quantization is applied to the query tokens, which are subsequently concatenated with the 2D mask tokens for reconstructing images at the decoder. 
The advantage of TiTok is its flexibility, as the number of query tokens can be controlled to balance the compression ratio and generation quality. 
GigaTok \cite{xiong2025gigatok} further improves the 1D tokenizer by scaling the model size with representation alignment \cite{yu2024representation}, which significantly enhances the generation quality of downstream AR models.

\noindent \textbf{Flexible Tokenizers.}
In contrast to fixed-grid tokenizers, flexible tokenizers aim to encode images with variable-length token sequences \cite{miwa2025one, bachmann2025flextok, yan2024elastictok, huang2025spectralar, liu2025detailflow, duggal2024adaptive, wang2025visual, esteves2025spectral, patel2026vibetoken, wen2025principal}. 
FlexTok \cite{bachmann2025flextok} applies nested dropout \cite{rippel2014learning} during tokenizer training, achieving coarse-to-fine image reconstruction from 1 to 256 tokens. 
However, FlexTok achieves the best gFID at 32 tokens, and generation quality decreases as more tokens are generated, indicating the under-utilization of tail tokens.
Meanwhile, One-D-Piece \cite{miwa2025one} adopts a similar dropout training strategy. 
However, it focuses on image reconstruction and lacks the analysis of the downstream autoregressive image generation.
Instead of representing the image at the original resolution with variable-length tokens, SpectralAR \cite{huang2025spectralar} and DetailFlow \cite{liu2025detailflow} introduce hierarchical reconstruction, where the early tokens reconstruct either low-resolution or low-frequency components of the image, while tail tokens complement fine-grained details. 
Consequently, using generated early tokens can only reconstruct low-resolution and blurry images, which undermines the flexibility of the AR models. 
Other flexible tokenizers, such as ViLex \cite{wang2025visual}, are designed for diffusion models, while ElasticTok \cite{yan2024elastictok} and ALIT \cite{duggal2024adaptive} mainly evaluate for image reconstruction.

\subsection{Representation Alignment for Generation}
Representation alignment \cite{yu2024representation} is initially designed for diffusion models \cite{peebles2023scalable, ma2024sit, wang2026persistence, fu2026frozen}, by aligning the diffusion models' intermediate features with vision foundation model features \cite{oquab2023dinov2, tschannen2025siglip}. 
Recently, VA-VAE \cite{yao2025reconstruction}, GigaTok \cite{xiong2025gigatok}, MAETok \cite{chen2025masked}, and ImageFolder \cite{li2024imagefolder} also inject semantic guidance from vision foundation models to regularize tokenizer training.  
Representation alignment significantly improves convergence speed and generation quality of the downstream diffusion and AR models. 

\section{Method}
\label{sec:method}

In this section, we first introduce the background of 1D and flexible tokenizers. 
We then discuss the generation bottleneck for flexible tokenizers using naive nested dropout.
Finally, we present ReTok, which incorporates proposed redundant token padding and hierarchical semantic regularization. 

\subsection{Preliminary}
\noindent \textbf{1D Discrete Tokenizers.} 1D discrete tokenizers, such as TiTok \cite{yu2024image}, aim to compress the image into the compact 1D sequence instead of the 2D grid.  
Given an image $\mathbf{X} \in\mathbb{R}^{H \times W \times 3}$, it is first patchified or fed into convolutional layers to obtain its patch embeddings $\mathbf{P} \in \mathbb{R}^{\frac{H}{f} \times \frac{W}{f} \times D}$, where $f$ denotes the downsampling ratio and $D$ is the embedding dimension. 
To construct the 1D image representations, a set of $N$ learnable query tokens $\mathbf{Q} \in \mathbb{R}^{N \times D}$ is initialized.   
A ViT encoder \cite{dosovitskiy2020image} then takes as input the concatenation of the query tokens $\mathbf{Q}$ and image embeddings $\mathbf{P}$.
The encoder only outputs query token embeddings, which are then quantized into the discrete tokens $\mathbf{Z} \in \mathbb{R}^{N \times d}$ using a quantizer:
\begin{equation}
\Zadj = \mathcal{Q}\!\left(\mathcal{E}\bigl([\mathbf{P};\,\mathbf{Q}]\bigr)\right),
\end{equation}
where $\mathcal{E}$ and $\mathcal{Q}$ are the encoder and quantizer, and $[\,\cdot;\cdot\,]$ denotes concatenation.
For reconstruction, learnable mask tokens $\mathbf{M} \in \mathbb{R}^{\frac{H}{f} \times \frac{W}{f} \times D}$ and quantized query tokens are fed to a ViT decoder $\mathcal{D}$ to recover the 2D image: $\hat{\mathbf{X}}=\mathcal{D}([\mathrm{MLP}(\mathbf{Z});~ \mathbf{M}])$.
% \begin{equation}
%     \hat{\mathbf{X}}=Dec(Proj(\mathbf{Z})\oplus \mathbf{M}).
% \end{equation}
$\mathbf{Z}$ is projected to embedding dimension $D$ by MLP.

\noindent \textbf{Flexible Tokenizers with Nested Dropout.} Previous works \cite{bachmann2025flextok, miwa2025one, liu2025detailflow, yan2024elastictok, huang2025spectralar} have proposed to apply nested dropout to train flexible tokenizers. 
During training, the tail tokens in the quantized token sequence $ \Zadj = [\mathbf{z}_1, \mathbf{z}_2, ..., \mathbf{z}_N]$ are randomly dropped, resulting in a truncated token sequence:
\begin{equation}
    \Zpadj=[\mathbf{z}_1, \mathbf{z}_2, ...,\mathbf{z}_{k}],
\end{equation}
where $k \le N$ denotes the number of tokens that are retained, while $N-k$ tail tokens are masked out.
In FlexTok \cite{bachmann2025flextok} and One-D-Piece \cite{miwa2025one}, $k$ is randomly sampled from $\{1, 2, 4, 8, ..., N\}$, while in DetailFlow \cite{liu2025detailflow}, $k$ is sampled from $\{8, 16, 24, ...,N\}$ with an interval of $8$.
The decoder aims to reconstruct the original image by using the truncated token sequence: 
\begin{equation}
    \hat{\mathbf{X}}=\mathcal{D}([\mathrm{MLP}(\mathbf{Z}');~ \mathbf{M}]).
\label{eq_drop}
\end{equation}
Using nested dropout, the tokenizer learns to encode images as ordered 1D coarse-to-fine sequences.  

\noindent \textbf{Training Tokenizers with Semantic Regularization.}
Existing works propose training visual tokenizers along with representation alignment \cite{xiong2025gigatok, yao2025reconstruction, yu2024representation, li2024imagefolder, chen2025masked} to improve the downstream generation performance. 
We follow GigaTok \cite{xiong2025gigatok}, which applies the semantic regularization to align the tokenizer's decoder features with DINOv2 \cite{oquab2023dinov2} image features from the same image:
\begin{equation}
    \mathcal{L}_{reg}= -\mathrm{cos}(\mathrm{MLP}(f_{l}^{\mathrm{dec}}), \mathbf{}f^{\mathrm{DINO}}),
\end{equation}
where $f_{l}^{\mathrm{dec}}$ denotes intermediate features from the tokenizer's $l$-th ViT decoder layer, $f^{\mathrm{DINO}}$ represents semantic features from the pre-trained DINOv2-B encoder on the same input image, and $\mathrm{MLP}$ projects $f_{l}^{\mathrm{dec}}$ to match the channel dimension of $f^{\mathrm{DINO}}$.
The full training objective of the tokenizer is the combination of the image reconstruction loss $\mathcal{L}_{rec}$ and the semantic regularization loss $\mathcal{L}_{reg}$:
\begin{equation}
    \mathcal{L}_{total} = \mathcal{L}_{rec} + \lambda\mathcal{L}_{reg}.
    \label{eq_loss}
\end{equation}
Here, we follow the reconstruction loss defined in VQGAN \cite{esser2021taming}, containing pixel-level reconstruction loss, perceptual loss \cite{johnson2016perceptual}, adversarial loss \cite{goodfellow2020generative,isola2017image}, and VQ codebook loss.
\begin{figure}
    \centering
    \includegraphics[width=\linewidth]{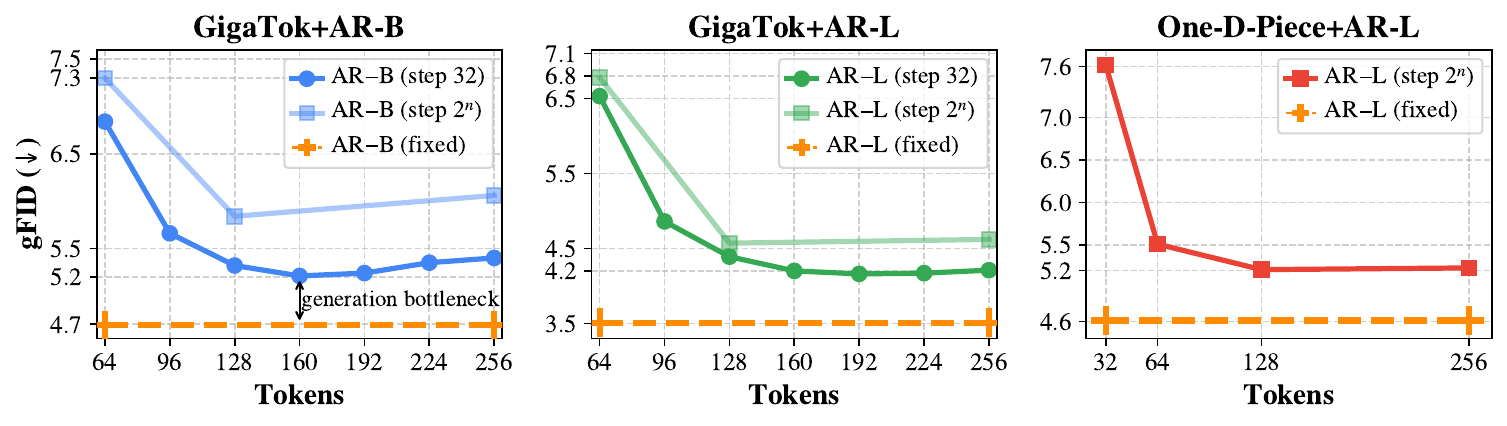}
    \caption{Illustration of the generation bottleneck in flexible tokenizers. We train AR models and evaluate generation quality at varying token lengths. Results indicate a significant decrease in downstream generation compared to fixed-length tokenizers. Details are in \cref{sec_finding}.}
    \label{fig2_bottleneck}
\end{figure}

\subsection{Generation Bottleneck for Flexible Tokenizers}
\label{sec_finding}
Even though there exist a few flexible tokenizers, limited work systematically analyzes their downstream autoregressive image generation performance, especially compared to the fixed-length counterparts.
We conduct our preliminary experiments based on GigaTok \cite{xiong2025gigatok}, a fixed-length 1D image tokenizer that achieves state-of-the-art image generation.
Following GigaTok, the token length $N$ is set to $256$, while we fine-tune GigaTok with $50$ epochs by using the naive nested dropout to obtain its flexible versions.
We conduct with two tokenizers, denoted as tokenizer (step $2^n$) and tokenizer (step 32), where the retained tokens $k$ are randomly selected from $\{32, 64, 128, 256\}$ and $\{32, 64, 96, ...,256\}$ with an interval of $32$, respectively. 
We then train the corresponding AR models\footnote{The AR model we utilize for evaluation is the LlamaGen-B (111M) and LlamaGen-L (343M) \cite{sun2024autoregressive}. GigaTok-S-S serves as the baseline tokenizer.} on ImageNet for $120$ epochs and evaluates the generation FID (gFID) on its validation set. 
Moreover, we evaluate One-D-Piece \cite{miwa2025one}, a pre-trained flexible tokenizer, and additionally train a version without nested dropout to serve as a fixed-length counterpart for comparison.
Our findings are as follows:

\noindent Finding 1. \textbf{Image generation quality decreases when flexible tokenizers are trained with the naive nested dropout.} We highlight the generation bottleneck of flexible tokenizers - AR models trained under flexible tokenizers perform worse than those trained on fixed-length tokenizers. 
In \cref{fig2_bottleneck}, we illustrate the results of different AR models under various token lengths.
Compared to the fixed-length baseline, all AR models with different parameters exhibit a significant decrease in generation quality. 
At the length $256$, AR-B (step $32$) and AR-B (step $2^n$) with GigaTok achieve $5.4$ and $6.1$ in gFID, respectively, whereas the fixed-length AR model reaches 4.69\footnote{We search the optimal CFG for AR models at the full length (256 tokens) and evaluate the gFID for the shorter length under the same CFG.}.
This performance drop underscores the limitations of naive nested dropout.

\noindent Finding 2. \textbf{Generating more tokens in the tail won't improve generation quality.}
As shown in \cref{fig2_bottleneck}, generating more tokens does not guarantee better quality and can even lead to degraded results; \eg, gFID of 256 tokens is worse than that at 160 tokens for AR-B (step 32) with GigaTok. 
Similar phenomena have been observed in AR-L.
We attribute this to naive nested dropout, where the tokenizer compresses most image information into early tokens.
This training strategy fails to exploit the tail tokens and impedes the further improvement of generation quality with increasing tokens.

\subsection{ReTok}
ReTok follows GigaTok's architecture \cite{xiong2025gigatok}.
Image patch embeddings $\mathbf{P}$ are obtained via a stack of convolution layers.
The ViT encoder and ViT decoder then learn the compact discrete 1D image representation. 
Finally, embeddings are mapped back to the pixel space by convolution layers for image reconstruction.
In the following part, we introduce key improvements for flexible tokenizers.

\noindent \textbf{Redundant Token Padding.} 
Training flexible tokenizers with naive nested dropout compresses most image information into early tokens, which incurs the generation bottleneck for autoregressive models. 
To overcome this problem, we propose redundant token padding by concatenating additional query tokens at the tail of the original sequence:
\begin{equation}
    [\mathbf{Z}; \mathbf{Z}_{pad}] = \mathcal{Q}(\mathcal{E}([\mathbf{P};\mathbf{Q};\mathbf{Q}_{pad}])),
\end{equation}
where $\mathbf{Q}_{pad}\in \mathbb{R}^{M\times D}$ denotes the $M$ padding tokens and $\mathbf{Z}_{pad}$ is the corresponding discrete tokens. 
The full token sequence $\mathbf{Z}_{full}$ is described as:
\begin{equation}
    [\mathbf{Z};\mathbf{Z}_{pad}] = [\mathbf{z}_1,...,\mathbf{z}_N, \mathbf{z}_{N+1}, ..., \mathbf{z}_{N+M}].
\end{equation}
We further perform nested dropout on the concatenated token sequence, where the number of retained tokens satisfies $k\le N+M$.
The truncated token sequence is fed to the decoder for reconstruction following \cref{eq_drop}.
During the tokenizer training, the original token sequence $\mathbf{Z}$ becomes the ``early" tokens in the current full token sequence, where the tokenizer aims to compress the most image information into it.
This training strategy activates the tail tokens in the original sequence. 
Since the original token sequence $\mathbf{Z}$ learns to represent the image during training, we solely use it and discard the encoded padding tokens $\mathbf{Z}_{pad}$ for downstream autoregressive generation. We illustrate the overview of our redundant token padding in \cref{fig_overview}.

\noindent \textbf{Hierarchical Semantic Regularization.} 
Previous works \cite{beyer2025highly} have shown that 1D tokenizers with high compression ratios (\eg, TiTok \cite{yu2024image} with 32 tokens) effectively learn semantic and high-level image representations. 
Inspired by their work, we propose hierarchical semantic regularization to enhance the semantic representation of early tokens while enabling the tail tokens for pixel-level reconstruction. 
We follow the training objective defined in \cref{eq_loss}, but make the regularization weight $\lambda$ a function of sequence length $k$:
\begin{equation}
    \mathcal{L}_{total} = \mathcal{L}_{rec} + \lambda(k) \mathcal{L}_{reg},
    \label{eq_our_loss}
\end{equation}
where $\lambda(k)$ decreases linearly as the retained sequence length $k$ increases. 
With \cref{eq_our_loss}, early tokens emphasize feature-level alignment with the semantic feature of DINOv2, while progressively enabling the subsequent tokens to complement the low-level structures and details of the image. 

\noindent \textbf{Decoder Fine-tuning.} 
Since early tokens are constrained with high semantic regularization during training, we further fine-tune the decoder to improve the reconstruction performance of early tokens. 
We freeze the well-trained encoder and the quantizer, while easing the semantic constraint by setting $\lambda(k)$ to a small constant for all sequence lengths. 
As we show in experiments, fine-tuning the decoder improves the quality of generated images for short token sequences (\eg, 32 or 64 tokens).

\begin{figure}[t]
    \centering

    % ================= 第一排：只放图片（使用 [b] 底部对齐） =================
    \begin{minipage}[t]{0.51\linewidth}
        \centering
        \includegraphics[width=\linewidth]{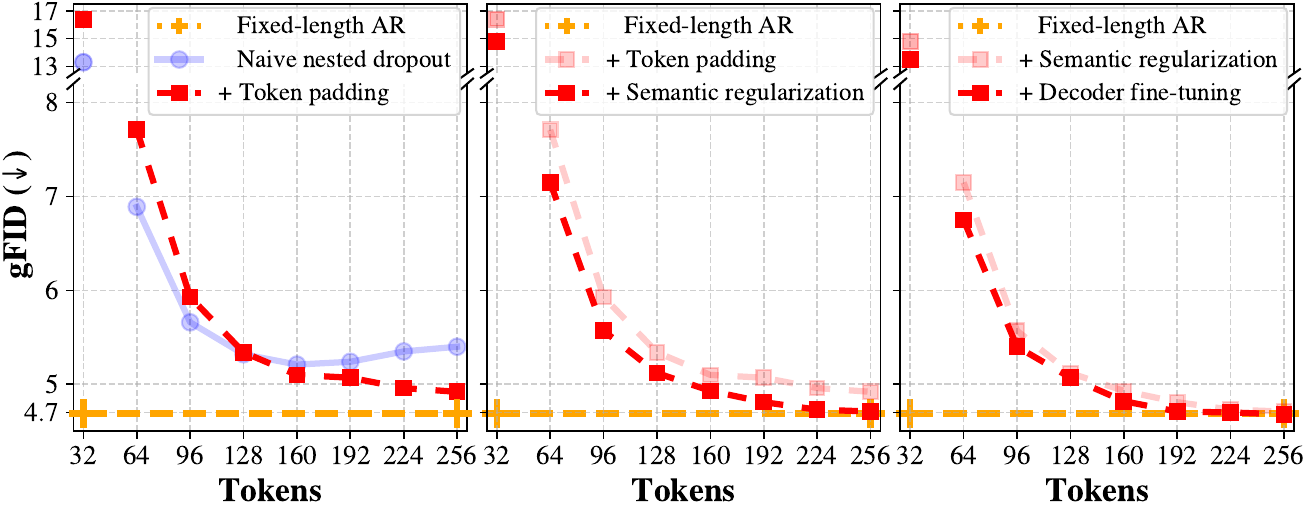}
    \end{minipage}
    \hfill 
    \begin{minipage}[t]{0.47\linewidth}
        \centering
        \includegraphics[width=0.99\linewidth]{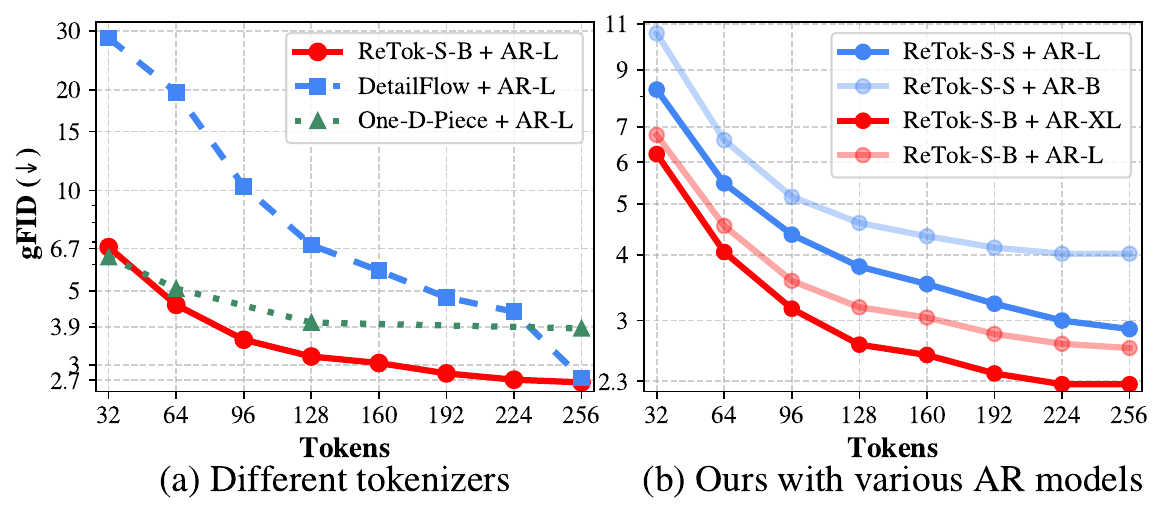}
    \end{minipage}\\ % <--- 核心：强制换行，彻底切断图片和Caption之间的垂直拉伸关联

    % ================= 第二排：只放 caption（使用 [t] 顶部对齐） =================
    \begin{minipage}[t]{0.51\linewidth}
        % 统一设置 skip 间距，确保 Caption 紧贴图片下方
        % \captionsetup{skip=4pt} 
        \caption{Ablation study on the improvement of ReTok. We evaluate the effectiveness of our proposed methods (redundant token padding, hierarchical semantic regularization, and decoder fine-tuning) on the ImageNet validation set.}
        \label{fig_roadmap}
    \end{minipage}
    \hfill
    \begin{minipage}[t]{0.47\linewidth}
        % \captionsetup{skip=4pt} 
        \caption{Analysis of image generation at different token lengths. (a) Downstream image generation comparison of ReTok, DetailFlow \cite{liu2025detailflow}, and One-D-Piece \cite{miwa2025one}. (b) Generation performance of ReTok with various AR models.}
        \label{fig_ours_gfid}
    \end{minipage}
    % \vspace{-3mm}
\end{figure}
\subsection{Application - Efficient Test-Time Scaling in AR Generation}
Since ReTok can decode images with varying lengths and consistently improves their quality as more tokens are generated, it naturally enables \textbf{efficient test-time scaling} (TTS) for \textbf{AR image generation}. 
For example, in a simple Best-of-$N$ (BoN) setting, early tokens in ReTok already produce images with decent quality and preserve the global structure of the final image.
Therefore, we can generate $N$ candidates using short prefixes (\eg, 64 tokens), select the best one with a verifier, and continue decoding the selected candidate, rather than generating $N$ full 256-token candidates, thereby reducing inference-time computation.
Detailed discussion is provided in \cref{sec_tts}.

\section{Experiments}
\label{sec_exp}
\subsection{Experiment Settings}
\label{sec_exp_setting}
\noindent \textbf{Implementation Details.} 
Following previous works \cite{bachmann2025flextok, liu2025detailflow, miwa2025one, xiong2025gigatok}, we set ReTok's original query token length $N$ to 256. 
The tokenizer can encode an image from 32 to 256 tokens, with step size 32.
We further extend the sequence length up to 480 by padding with 224 redundant query tokens, which enables the number of retained tokens $k$ to be randomly selected from $\{32, 64, ...,256,288,...,480\}$ when applying nested dropout during tokenizer training.
This keeps the ``average'' retained sequence length around 256.
Following the architecture design of GigaTok \cite{xiong2025gigatok}, we train two versions of the tokenizer, ReTok-S-S (136M) and ReTok-S-B (232M), with a codebook size of 16384.
Unless otherwise specified, we initialize our tokenizers with pre-trained GigaTok weights. We train the ReTok-S-S for 200 epochs and ReTok-S-B for 250 epochs, while all decoders of tokenizers are further fine-tuned for 50 epochs.
The weight of semantic regularization $\lambda(k)$ for ReTok-S-S and ReTok-S-B in \cref{eq_our_loss} decreases linearly from 2.0 and 2.5 to 0.5, respectively, for sequence lengths between 32 and 256, and is set to 0.5 for lengths beyond 256. 
For fine-tuning the decoder, we fix the semantic constraint to 0.5.

For downstream image generation, we discard the redundant discrete tokens and keep the token length to 256.  
Our autoregressive models are based on LlamaGen \cite{sun2024autoregressive}. For the ReTok-S-S and ReTok-S-B, we train the LlamaGen-B (111M)/LlamaGen-L (343M) and LlamaGen-L (343M)/LlamaGen-XL (775M) variants, respectively. 
The AR models in our main experiments are trained for 300 epochs, following the training recipes defined in \cite{sun2024autoregressive}.
During inference, a step-function Classifier-Free Guidance (CFG) schedule is employed, where the first $18\%$ of tokens are generated without CFG (CFG = 1) to enhance diversity, and the remaining tokens use CFG to improve generation quality.
We search for the optimal CFG for each AR model during evaluation.
All tokenizers and AR models are trained on ImageNet \cite{russakovsky2015imagenet} with images of size $256 \times 256$, and evaluated on the ImageNet validation set.

\noindent \textbf{Metrics.} We apply Fréchet Inception Distance (FID) \cite{heusel2017gans}, Inception Score (IS) \cite{salimans2016improved}, Precision and Recall \cite{kynkaanniemi2019improved} to evaluate the image generation performance. 
We also report Peak Signal-to-Noise Ratio (PSNR), Structural Similarity Index (SSIM) \cite{wang2004image}, Learned Perceptual Image Patch Similarity (LPIPS) \cite{lpips}, and reconstruction FID (rFID) results for assessing the tokenizer's reconstruction performance.

\subsection{Roadmap to Improve ReTok}
\label{sec_exp_roadmap}
We evaluate each part of the proposed method. (1) We start from GigaTok-S-S and train the fixed-length AR model (111M) for 120 epochs. 
The AR model achieves a gFID of 4.69, serving as our reference.
(2) Building upon this baseline, we train the flexible tokenizer with naive nested dropout for 50 epochs, where the retained sequence length $k$ is selected from $\{32, 64, 96, ...,256\}$.
The downstream AR model uses the same configuration as above. 
We search for the optimal CFG at full length (\eg, 256 tokens) and apply it to shorter sequences. 
The training objective follows \cref{eq_loss} for all sequence lengths with $\lambda=0.5$. 
(3) In \cref{fig_roadmap} left, the AR model achieves the best gFID of 5.21 at 160 tokens and a worse gFID of 5.40 at 256 tokens, showing a significant decrease compared to the fixed-length counterpart.
(4) We then adopt the proposed redundant token padding strategy to train the flexible tokenizer. \cref{fig_roadmap} left shows that the downstream AR model achieves the best gFID of 4.92 at 256 tokens. 
(5) However, generation quality decreases at early tokens. We therefore propose hierarchical semantic regularization to enhance their semantic representation. 
This technique improves gFID of 32 tokens by 1.61 while attaining a gFID of 4.71 at 256 tokens, comparable to the baseline. 
(6) Finally, we relax the semantic constraint and fine-tune the decoder for better image reconstruction. \cref{fig_roadmap} right shows that decoder fine-tuning refines image generation at short token lengths while maintaining quality for long sequences.

\begin{table}[t!]
\centering
\setlength{\tabcolsep}{6pt}
\renewcommand{\arraystretch}{1.}
\resizebox{\textwidth}{!}{
\begin{tabular}{llcc | lccccccc}
\toprule
Type & Tokenizer & Param. & rFID$\downarrow$ & Generator & Param. & Type & \# Tokens & gFID$\downarrow$ & gIS$\uparrow$ & Precision$\uparrow$ & Recall$\uparrow$ \\
\cmidrule(lr){1-12}
% 这一行用了 multicolumn{12}{c}，这里的 c 没有竖线，所以这一行中间是空的，符合你的要求
\multicolumn{12}{c}{\textit{\textbf{Continuous modeling}}}\\
\cmidrule(lr){1-12}

% ----- SD-VAE -----
% 下面的行会自动带上竖线
\multirow{3}{*}{2D} & \multirow{3}{*}{SD-VAE~\cite{rombach2022high}} & \multirow{3}{*}{84M} & \multirow{3}{*}{0.62} 
& DiT-XL/2~\cite{peebles2023scalable} & 675M & Diff & 1024 & 2.27 & 278.2 & 0.83 & 0.57 \\
& & & & SiT-XL/2~\cite{ma2024sit} & 675M & Diff & 1024 & 2.06 & 270.3 & 0.82 & 0.59 \\
& & & & SiT-XL/2+REPA~\cite{yu2024representation}& 675M & Diff & 1024 & 1.42 & 305.7 & 0.80 & 0.65 \\

2D & VA-VAE~\cite{yao2025reconstruction} & 70M & 0.28 & LightningDiT~\cite{yao2025reconstruction} & 675M & Diff & 256 & 1.35 & 295.3 & 0.79 & 0.65 \\
2D & VAE~\cite{li2024autoregressive} & 66M & 0.53 & MAR-B~\cite{li2024autoregressive} & 208M & AR+Diff & 256 & 2.31 & 281.7 & 0.82 & 0.57 \\

\cmidrule(lr){1-12}
\multicolumn{12}{c}{\textit{\textbf{Discrete modeling}}}\\
\cmidrule(lr){1-12}
2D & VQGAN~\cite{chang2022maskgit} & 66M & 2.28 & MaskGIT~\cite{chang2022maskgit} & 227M & Mask & 256 & 6.18 & 182.1 & 0.80 & 0.51 \\
\multirow{2}{*}{2D} & \multirow{2}{*}{VAR~\cite{tian2024visual}} & \multirow{2}{*}{109M} & \multirow{2}{*}{0.90}
& VAR-d16~\cite{tian2024visual} & 310M & VAR & 680 & 3.30 & 274.1 & 0.84 & 0.51 \\
& & & & VAR-d20~\cite{tian2024visual} & 600M & VAR & 680 & 2.57 & 302.6 & 0.83 & 0.56 \\
2D & ImageFolder~\cite{li2024imagefolder} & 176M & 0.80 & ImageFolder-VAR~\cite{li2024imagefolder} & 362M & VAR & 286 & 2.60 & 295.0 & 0.75 & 0.63 \\
2D & LlamaGen~\cite{sun2024autoregressive} & 72M & 2.19 & LlamaGen-L~\cite{sun2024autoregressive} & 343M & AR & 256 & 3.81 & 248.3 & 0.83 & 0.52 \\

1D & VFMTok$^\dagger$~\cite{zheng2025vision} & --  & 0.89 & LlamaGen-L$^\dagger$~\cite{sun2024autoregressive} & 343M & AR  & 256 & 2.11 & 230.1 & 0.82 & 0.60 \\
1D & VFMTok~\cite{zheng2025vision} & --  & 1.02 & LlamaGen-L~\cite{sun2024autoregressive} & 343M & AR  & 256 & 2.79 & 276.0   & --   & --   \\
1D & TiTok-S~\cite{yu2024image} & 72M  & 1.71 & MaskGIT-UViT-L~\cite{chang2022maskgit} & 287M & Mask & 128 & 1.87 & 281.8 & --   & --   \\
1D & TiTok-L~\cite{yu2024image} & 641M & 2.21 & MaskGIT-ViT~\cite{chang2022maskgit}    & 177M & Mask &  32 & 2.77 & 199.8 & --   & --   \\
\multirow{2}{*}{1D} & \multirow{2}{*}{GigaTok-S-S~\cite{xiong2025gigatok}} & \multirow{2}{*}{136M} & \multirow{2}{*}{1.01}
& LlamaGen-B~\cite{sun2024autoregressive} & 111M & AR & 256 & 4.05 & 240.6 & 0.81 & 0.51 \\
& & & & LlamaGen-L$^\star$~\cite{sun2024autoregressive} & 343M & AR & 256 & 2.86 & 261.2 & 0.81 & 0.57 \\
1D & GigaTok-S-B \cite{xiong2025gigatok} & 232M & 0.89 & LlamaGen-L$^\star$ \cite{sun2024autoregressive} & 343M & AR & 256 & 2.71 & 246.3 & 0.81 & 0.58
\\
\cmidrule(lr){1-12}
\multirow{2}{*}{Flex} & \multirow{2}{*}{FlexTok~\cite{bachmann2025flextok}} & \multirow{2}{*}{$\sim 2.5$B} & \multirow{2}{*}{1.08} & \multirow{2}{*}{LlamaGen~\cite{bachmann2025flextok}} & \multirow{2}{*}{1.33B} & \multirow{2}{*}{AR+Diff} & 32 & \textbf{1.86} & -- & -- & --\\
& & & & & & & 256 & $\sim 2.5$ & -- & -- & --
\\
\multirow{2}{*}{Flex} & \multirow{2}{*}{One-D-Piece~\cite{miwa2025one}} & \multirow{2}{*}{641M} & \multirow{2}{*}{1.08} & LlamaGen-B$^\star$~\cite{sun2024autoregressive} & 86M & AR & 256 & 6.49 & 194.3 & \textbf{0.82} & 0.43 \\
& & & & LlamaGen-L$^\star$~\cite{sun2024autoregressive} & 318M & AR & 256 & 3.86 & 231.7 & 0.81 & 0.51\\
Flex & DetailFlow-32~\cite{liu2025detailflow} & 270M & 0.80 & LlamaGen-L~\cite{sun2024autoregressive} & 326M & AR & 256 &  2.75 & 250.8 & 0.81 & 0.58 \\
Flex & DetailFlow-64~\cite{liu2025detailflow} & 270M & 0.55 & LlamaGen-L~\cite{sun2024autoregressive} & 326M & AR & 512 &  2.62 & 245.3 & 0.80 & \textbf{0.60} \\
\multirow{2}{*}{Flex} & \multirow{2}{*}{SpectralAR~\cite{huang2025spectralar}} & \multirow{2}{*}{--} & \multirow{2}{*}{4.03} & SpectralAR-d16~\cite{huang2025spectralar} & 310M & AR & 64 & 3.02 & \underline{282.2} & 0.81 & 0.55 \\
& & & & SpectralAR-d20~\cite{huang2025spectralar} & 600M & AR & 64 & 2.49 & \textbf{305.4} & -- & --
\\
\cmidrule(lr){1-12}
\multirow{2}{*}{Flex} & \multirow{2}{*}{\textbf{ReTok-S-S}} & \multirow{2}{*}{136M} & \multirow{2}{*}{1.09} & LlamaGen-B~\cite{sun2024autoregressive} & 111M & AR & 256 & 4.02 & 245.2 & 0.80 & 0.50 \\
& & & & LlamaGen-L~\cite{sun2024autoregressive} & 343M & AR & 256 & 2.92 & 243.5 & 0.81 & 0.57 \\
\multirow{2}{*}{Flex} & \multirow{2}{*}{\textbf{ReTok-S-B}} & \multirow{2}{*}{232M} & \multirow{2}{*}{1.01} & LlamaGen-L~\cite{sun2024autoregressive} & 343M & AR & 256 & 2.66 & 231.7 & \textbf{0.82} & 0.57 \\
& & & & LlamaGen-XL~\cite{sun2024autoregressive} & 775M & AR & 256 & \underline{2.27} & 245.9 & \textbf{0.82} & \textbf{0.60} \\
\bottomrule
\end{tabular}
}
\caption{Comparison of class-conditional image generation on ImageNet $256\times256$. $\dagger$ denotes the model generates images at $336 \times 336$ resolution, and are resized to $256\times256$ for evaluation. $\star$ indicates models that are our implementation. We report the optimal gFID achieved across scenarios with and without Classifier-Free Guidance. \textbf{Bold} and \underline{underline} indicate the first and second best methods within flexible tokenizers.}
% \vspace{-3mm}
\label{tab_quan_results}
\end{table}

\subsection{Main Results}

\begin{figure}[t!]
\centering
\begin{spacing}{0.20} % 设置行间距
\setlength{\tabcolsep}{0.5pt} % 设置列间距
\begin{tabular}{cccccc}

        \begin{minipage}[c]{0.16\linewidth}
        \centering
        \includegraphics[width=\linewidth]{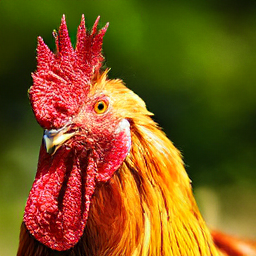}\\[2pt]
        % \captionof{figure}{(a)}
    \end{minipage}
    &
    \begin{minipage}[c]{0.16\linewidth}
        \centering
        \includegraphics[width=\linewidth]{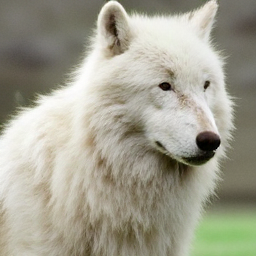}\\[2pt]
    \end{minipage} &
    
    \begin{minipage}[c]{0.16\linewidth}
        \centering
        \includegraphics[width=\linewidth]{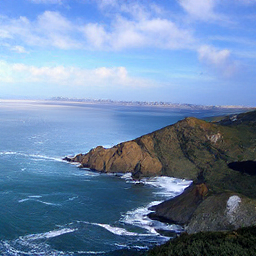}\\[2pt]
    \end{minipage} &
    
    \begin{minipage}[c]{0.16\linewidth}
        \centering
        \includegraphics[width=\linewidth]{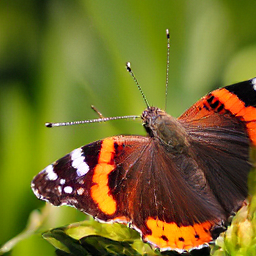}\\[2pt]
    \end{minipage}  &
    
    \begin{minipage}[c]{0.16\linewidth}
        \centering
        \includegraphics[width=\linewidth]{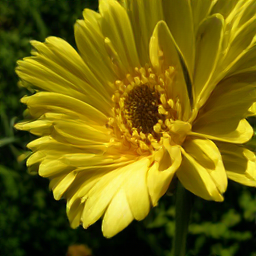}\\[2pt]
    \end{minipage} &

    \begin{minipage}[c]{0.16\linewidth}
        \centering
        \includegraphics[width=\linewidth]{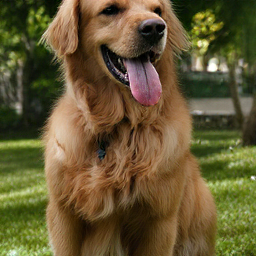}\\[2pt]
    \end{minipage} 
     \\
    
    \begin{minipage}[c]{0.16\linewidth}
        \centering
        \includegraphics[width=\linewidth]{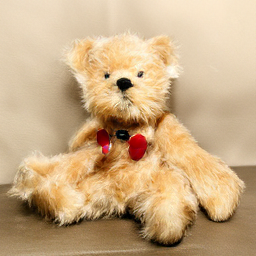}\\[2pt]
    \end{minipage} &

    \begin{minipage}[c]{0.16\linewidth}
        \centering
        \includegraphics[width=\linewidth]{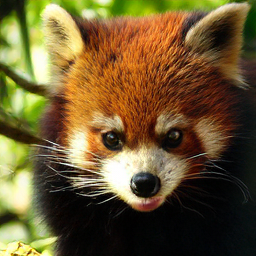}\\[2pt]
    \end{minipage} &

    \begin{minipage}[c]{0.16\linewidth}
        \centering
        \includegraphics[width=\linewidth]{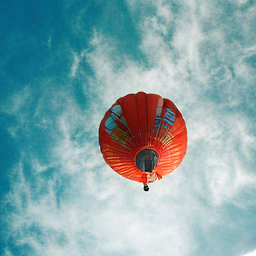}\\[2pt]
    \end{minipage} &

    \begin{minipage}[c]{0.16\linewidth}
        \centering
        \includegraphics[width=\linewidth]{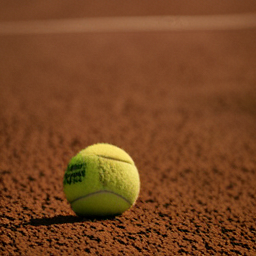}\\[2pt]
    \end{minipage} &

    \begin{minipage}[c]{0.16\linewidth}
        \centering
        \includegraphics[width=\linewidth]{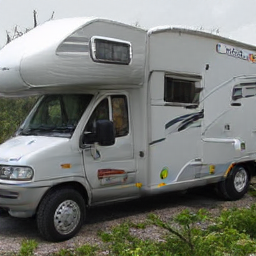}\\[2pt]
    \end{minipage} &

    \begin{minipage}[c]{0.16\linewidth}
        \centering
        \includegraphics[width=\linewidth]{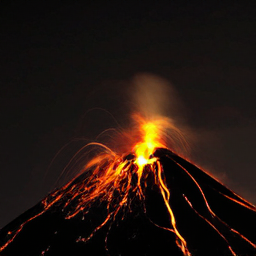}\\[2pt]
    \end{minipage} 

    \\
    \begin{minipage}[c]{0.16\linewidth}
        \centering
        \includegraphics[width=\linewidth]{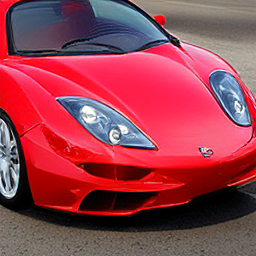}\\[2pt]
    \end{minipage} &

    \begin{minipage}[c]{0.16\linewidth}
        \centering
        \includegraphics[width=\linewidth]{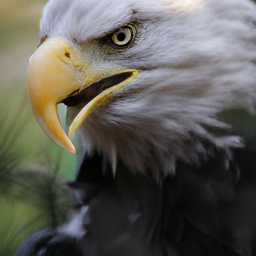}\\[2pt]
    \end{minipage} &

    \begin{minipage}[c]{0.16\linewidth}
        \centering
        \includegraphics[width=\linewidth]{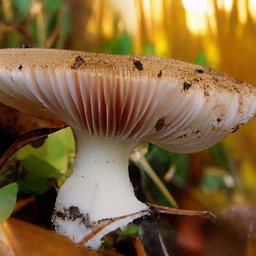}\\[2pt]
    \end{minipage} &

    \begin{minipage}[c]{0.16\linewidth}
        \centering
        \includegraphics[width=\linewidth]{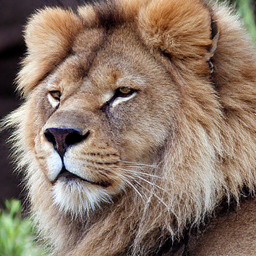}\\[2pt]
    \end{minipage} &

    \begin{minipage}[c]{0.16\linewidth}
        \centering
        \includegraphics[width=\linewidth]{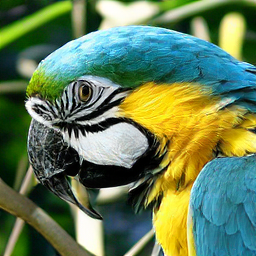}\\[2pt]
    \end{minipage} &

    \begin{minipage}[c]{0.16\linewidth}
        \centering
        \includegraphics[width=\linewidth]{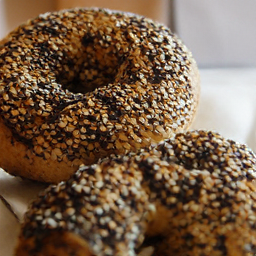}\\[2pt]
    \end{minipage} 
    \\[2pt]
\end{tabular}
\end{spacing}
\vspace{1mm}
\caption{Examples of generated images on ImageNet $256\times256$ from the ReTok-S-B with LlamaGen-XL models using 256 tokens. The classifier-free guidance is set to $4.0$.
}
\label{fig_visual_examples}
\end{figure}

\noindent \textbf{Class-conditional Image Generation.} 
We first evaluate the image generation performance at the full length on ImageNet. 
As shown in \cref{tab_quan_results}, our ReTok achieves performance comparable to that of our GigaTok baseline \cite{xiong2025gigatok}, demonstrating its effectiveness in addressing the generation bottleneck for flexible tokenizers. 
Compared to other flexible tokenizers, our method presents a significant gain over One-D-Piece \cite{miwa2025one} on downstream image generation.
Moreover, ReTok outperforms DetailFlow \cite{liu2025detailflow} under similar parameters.
%
% For instance, ReTok-S-B with LlamaGen-L (343M) attains gFID of 2.66, which is lower than the gFID of 2.75 obtained by DetailFlow-32 with LlamaGen (326M).
%
% For FlexTok \cite{bachmann2025flextok}, it fails to consistently improve the generation quality by generating more tokens; it achieves a gFID of 1.86 at 32 tokens and about a gFID of 2.5 at 256.
% %
% Note that FlexTok employs a large tokenizer and generator in conjunction with a diffusion decoder. 
%
\cref{fig_visual_examples} shows several generated images by using 256 tokens.

We further compare the generation results of different flexible tokenizers using the same LlamaGen-L across various token lengths.
For DetailFlow \cite{liu2025detailflow}, since it generates images at varying resolutions for different token lengths, we resized all low-resolution outputs to 256$\times$256 before measuring gFID.
\cref{fig_ours_gfid} (a) presents that our method outperforms the One-D-Piece across all token lengths except at 32 tokens. 
Notably, One-D-Piece shows negligible gains from 128 to 256 tokens.
In contrast, our method improves the gFID from 3.18 to 2.66 between 128 and 256 tokens.
%
% Compared to DetailFlow, it performs similar generation quality at 256 tokens but deteriorates rapidly at shorter lengths.
%
We also evaluate the generation quality of our method with different AR models in \cref{fig_ours_gfid} (b).
The results reveal consistent improvements with the increasing number of tokens. 
%
% Visual examples of progressive generation of our method can be found in \cref{fig_overview} (c) and supplementary material.

\noindent \textbf{Image Reconstruction.} 
We demonstrate the image reconstruction quality of our method in \cref{tab_quan_results}, where our ReTok-S-S and ReTok-S-B achieve rFID scores of 1.09 and 1.01 with 256 tokens, respectively.
Our tokenizers slightly decrease compared to GigaTok, which we attribute to the use of nested dropout.
Nevertheless, ReTok-S-B outperforms methods One-D-Piece and FlexTok, demonstrating its overall effectiveness in reconstruction. Detailed reconstruction results can be found in the appendix.
%
% We present reconstructed images in \cref{fig_reconstruction}, where the results show that our method is capable of recovering plausible images at both 32 and 256 token lengths. 

\subsection{Results of Test-Time Scaling}
\label{sec_tts}
\begin{figure}[t]
    \centering

    % ================= 第一排：只放图片（使用 [b] 底部对齐） =================
    \begin{minipage}[c]{0.48\linewidth}
        \centering
        \includegraphics[width=\linewidth]{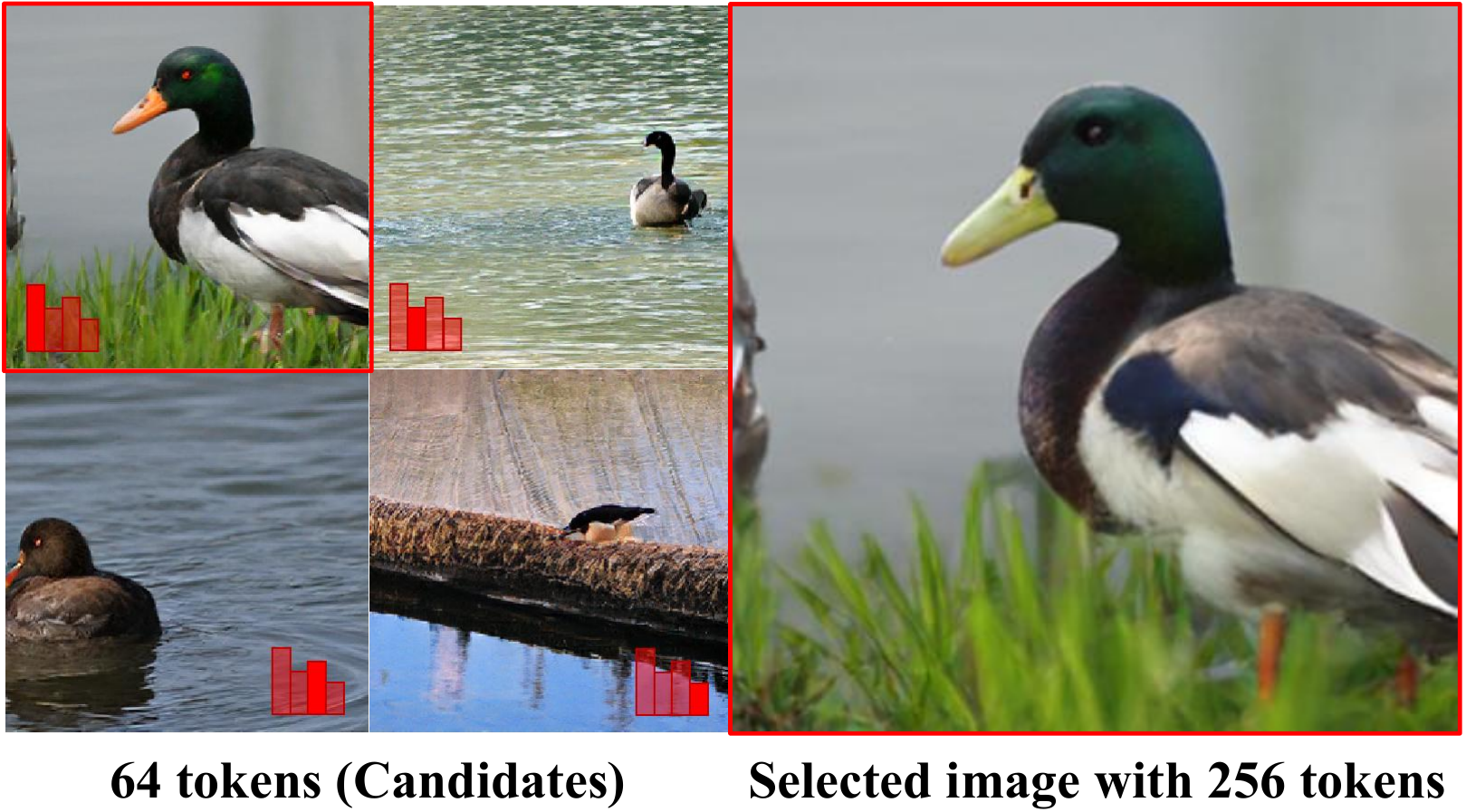}
    \end{minipage}
    \hfill 
    \begin{minipage}[c]{0.48\linewidth}
        \centering
        \includegraphics[width=\linewidth]{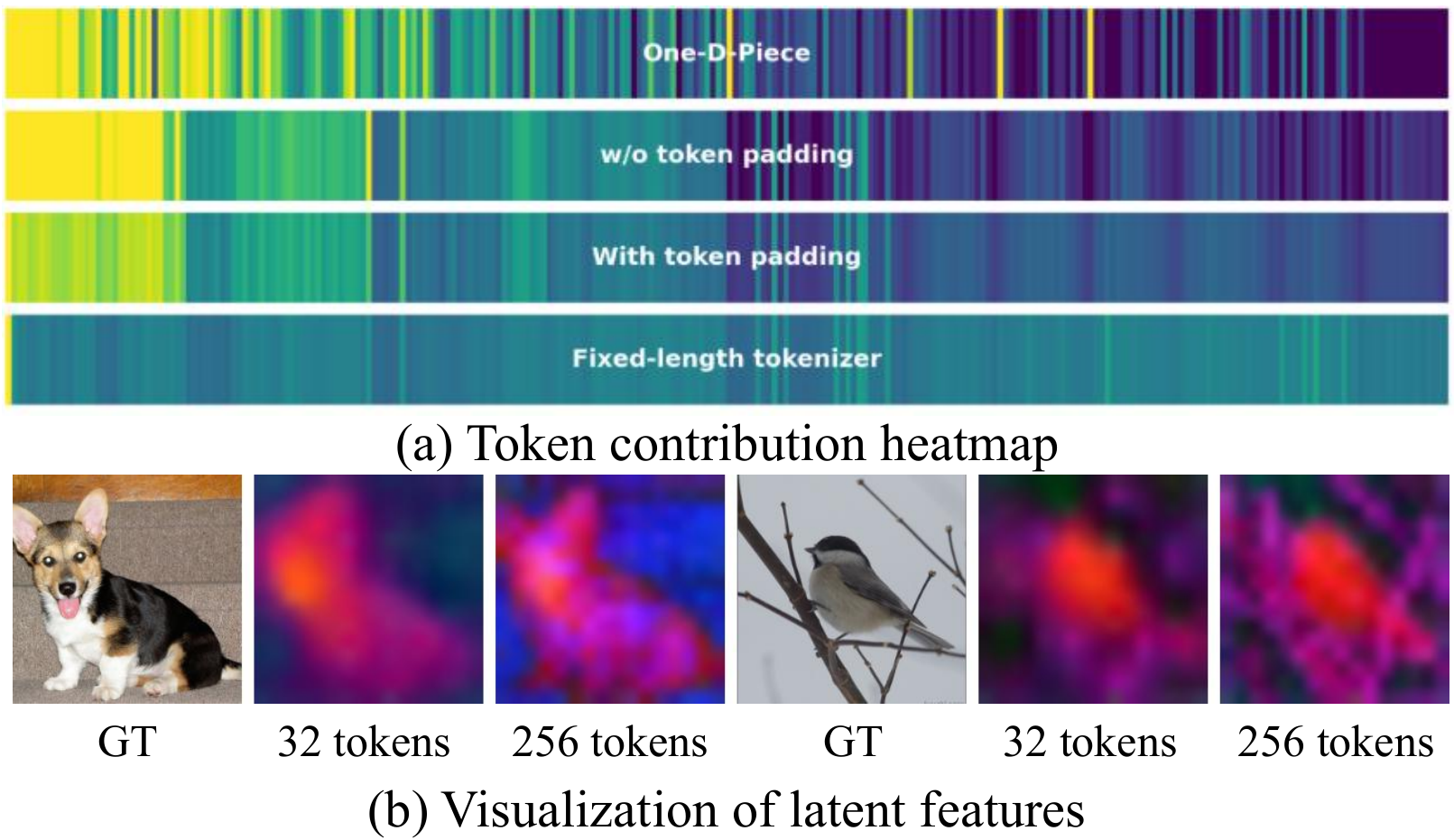}
    \end{minipage}\\ % <--- 核心：强制换行，彻底切断图片和Caption之间的垂直拉伸关联

    % ================= 第二排：只放 caption（使用 [t] 顶部对齐） =================
    \begin{minipage}[t]{0.48\linewidth}
        % 统一设置 skip 间距，确保 Caption 紧贴图片下方
        % \captionsetup{skip=4pt} 
        \caption{Visualizing Test-Time Scaling (TTS) in AR generation. Candidates (64 tokens) are first generated and verified, where the selected image is used to further decode up to 256 tokens.}
        \label{fig_tts}
    \end{minipage}
    \hfill
    \begin{minipage}[t]{0.48\linewidth}
        % \captionsetup{skip=4pt} 
        \caption{Ablation study of ReTok. (a) Analysis of token contribution (S-S tokenizer). (b) Visualization of latent features from the tokenizer decoder at different token lengths.}
        \label{fig_heatmap}
    \end{minipage}
% \vspace{-10mm}
\end{figure}

To verify test-time scaling (TTS) in ReTok, we conduct experiments on AR-B and AR-L models trained with ReTok-S-S (CFG=1). Following \cite{ma2025inference}, we implement Best-of-$N$ (BoN) selection. We generate 64 early tokens and reconstruct them for verification. 
To select the best candidate, we use the pre-trained linear classification head of DINOv2 \cite{oquab2023dinov2} and choose the candidate with the highest logit for the target class.
The selected image is used to decode subsequent tokens (see \cref{fig_tts}).
% \begin{figure}[t]
%     \centering
%     \includegraphics[width=\linewidth]{Figs/fig_reconstruction_eccv.pdf}
%     \caption{Image reconstruction comparison with various tokens. Low-resolution results of DetailFlow \cite{liu2025detailflow} are resized to 256$\times$256. Our tokenizer shows high-fidelity reconstruction at 32 and 256 tokens.}
%     \label{fig_reconstruction}
% \end{figure}
\begin{table}[t]
    \centering

    % ================= 第一排：图表内容（垂直居中对齐 [c]） =================
    \begin{minipage}[c]{0.345\linewidth}
        \centering
        \resizebox{\linewidth}{!}{
            \begin{tabular}{@{}lcccc@{}}
                \toprule 
                Sem. Reg. $\lambda(k)$ & Tokens & rFID\sm{$\downarrow$} & LPIPS\sm{$\downarrow$} & gFID\sm{$\downarrow$} \\
                \midrule
                \multirow{2}{*}{0.5} & 32 & 9.92 & 0.411 & 16.40 \\
                & 256 & 1.15 & 0.232 & 4.92 \\
                \midrule
                \multirow{2}{*}{2.0--0.5} & 32 & 9.02 & 0.415 & 14.79 \\
                & 256 & 1.18 & 0.234 & 4.68\\
                \midrule
                2.0 & 256 & 1.24 & 0.241 & 4.72 \\
                \bottomrule
            \end{tabular}
        }
    \end{minipage}
    \hfill
    \begin{minipage}[c]{0.625\linewidth}
        \centering
        \resizebox{\linewidth}{!}{
            \begin{tabular}{@{}lcccccc@{}}
                \toprule
                Method & Search size & FID\sm{$\downarrow$} & IS\sm{$\uparrow$}& Precision\sm{$\uparrow$}& Recall\sm{$\uparrow$} & CLIP score\sm{$\uparrow$}\\
                \midrule
                \color{gray} AR-B & \color{gray} 1 & \color{gray} 8.46 & \color{gray} 105.6 & \color{gray} 0.74 & \color{gray} 0.59 & \color{gray} 28.30 \\
                AR-B & 4 & 4.69 & 194.0 & 0.83 & 0.51 & 29.59 \\
                \color{gray} AR-L & \color{gray} 1 & \color{gray} 3.99 & \color{gray} 159.3 & \color{gray} 0.77 & \color{gray} 0.61 & \color{gray} 28.84 \\
                AR-L & 4 & 3.67 & 280.4 & 0.85 & 0.53 & 29.91 \\
                \bottomrule
            \end{tabular}
        }
    \end{minipage}\\ 
    
    \begin{minipage}[t]{0.35\linewidth} 
        \captionsetup{font=small, skip=4pt, position=bottom} 
        \caption{Ablation of semantic regularization. The hierarchical regularization improves generation quality while maintaining decent reconstruction.}
        \label{tab2_semantic_reg}
    \end{minipage}
    \hfill
    \begin{minipage}[t]{0.60\linewidth} 
        \captionsetup{font=small, skip=4pt, position=bottom} 
        \caption{Quantitative results of Test-Time Scaling (TTS) on ReTok-S-S. We compare the generation performance without TTS (search size = 1) and with TTS (search size = 4). The CFG is set to 1 during the generation process.}
        \label{tab3_tts_table}
    \end{minipage}
\end{table}

Experimental results in \cref{tab3_tts_table} show that TTS improves generation quality across FID, IS, and CLIP scores \cite{radford2021learning} (using prompt: \textit{"a photo of \{class label\}"}) at search size $N=4$. However, we find recall decreases, which suggests TTS trades sample diversity for semantic consistency, consistent with \cite{ma2025inference}. 
%
% In \cref{fig_tts}, we provide visual examples of TTS. Candidate images with 64 tokens already exhibit sufficient semantic structure for verification. The selected image is further decoded to 256 tokens, where local textures are significantly enhanced.

\subsection{More Analysis and Ablation Study}
\noindent \textbf{Redundant Token Padding is the Key to Activate Tail Tokens.} 
As illustrated in \cref{fig_roadmap}, token padding improves overall generation quality as more tail tokens are generated. 
We further analyze token contribution following One-D-Piece \cite{miwa2025one}. 
We first reconstruct an image $\hat{\mathbf{X}}$ with the tokenizer, and measure a token's contribution by the L1 distance $\|\hat{\mathbf{X}}-\hat{\mathbf{X}}^\prime\|$ between the original reconstruction and perturbed version $\hat{\mathbf{X}}^\prime$, where token $\mathbf{z}_i$ at position $i$ is replaced by a random token. 
We compute the mean L1 distance for each position $i$ over the ImageNet validation set, then apply softmax to obtain a normalized contribution distribution.
We visualize the token contribution heatmap in \cref{fig_heatmap} (a), comparing One-D-Piece, ReTok without token padding, ReTok with token padding, and GigaTok.  
As expected, both One-D-Piece and ReTok without token padding show highly concentrated contributions at head tokens (yellow), indicating their dominant role in reconstruction, while tail tokens contribute negligibly.
In contrast, our method effectively activates tail tokens, yielding a more uniform contribution distribution across all tokens, similar to the fixed-length tokenizer.

\noindent \textbf{Early Tokens Deserve High Semantic Regularization.}
We justify the proposed hierarchical semantic regularization in \cref{tab2_semantic_reg}. 
We first compare our tokenizer trained with hierarchical semantic regularization against a baseline without it, where the baseline regularization weight is set to a low constant (0.5) for all tokens (see details in \cref{sec_exp_roadmap}). 
Hierarchical semantic regularization significantly improves overall generation quality across all token lengths without compromising reconstruction performance.
Furthermore, we consider an extreme case by setting a large fixed regularization weight of 2.0, identical to the weight applied at 32 tokens in the hierarchical version. However, the results show severe degradation in reconstruction and no improvement in generation.
\begin{table}[t!]
    \centering
    \begin{minipage}[b]{0.35\linewidth}
        \centering
        \resizebox{\linewidth}{!}{%
            \begin{tabular}[b]{cccc}
             \toprule 
             Pad. Tokens & rFID$\downarrow$ & LPIPS$\downarrow$ & gFID$\downarrow$ \\
             \midrule
             64  & 1.21 & 0.238 & 4.74 \\
             224 & 1.18 & 0.234 & 4.68 \\
             384 & 1.34 & 0.245 & 5.03 \\
             \bottomrule
            \end{tabular}%
        }
    \end{minipage}\hfill
    \begin{minipage}[b]{0.282\linewidth}
        \centering
        \resizebox{\linewidth}{!}{%
            \begin{tabular}[b]{cccc}
             \toprule 
             $\lambda_{\text{start}}$ & rFID$\downarrow$ & LPIPS$\downarrow$ & gFID$\downarrow$ \\
             \midrule
             5 & 1.27 & 0.242 & 4.93\\
             2 & 1.18 & 0.234 & 4.68\\
             1 & 1.13 & 0.232 & 4.75 \\
             \bottomrule
            \end{tabular}%
        }
    \end{minipage}\hfill
    \begin{minipage}[b]{0.309\linewidth} 
        \centering
        \resizebox{\linewidth}{!}{%
            \begin{tabular}[b]{ccccc}
             \toprule 
             Step Size & Tokens & rFID$\downarrow$ & LPIPS$\downarrow$ & gFID$\downarrow$ \\
             \midrule
             \multirow{2}{*}{16} & 32 & 9.56 & 0.422 & 15.79\\
             & 256 & 1.22 & 0.234 & 4.71\\
             \midrule
             \multirow{2}{*}{32} & 32 & 9.02 & 0.415 & 14.79 \\
             & 256 & 1.18 & 0.234 & 4.68 \\
             \bottomrule
            \end{tabular}%
        }
    \end{minipage}

    \medskip 
    \begin{minipage}[t]{0.35\linewidth}
        \centering
        {\footnotesize (a) Number of padding tokens}
    \end{minipage}\hfill
    \begin{minipage}[t]{0.28\linewidth}
        \centering
        {\footnotesize (b) Initial weight $\lambda_{\text{start}}$}
    \end{minipage}\hfill
    \begin{minipage}[t]{0.33\linewidth}
        \centering
        {\footnotesize (c) Dropout step size}
    \end{minipage}

    \medskip

    % ================= 总标题 =================
    \caption{Design choices of ReTok (S-S tokenizer). (a) Effect of the number of padding tokens. The reconstruction and generation performance are evaluated on the full token (256) length. (b) Effect of the initial weight of the hierarchical semantic regularization. We vary the semantic weight $\lambda(32)$ while keeping $\lambda(256)$ at 0.5. (c) Impact of the step size of nested dropout.}
    \label{tab4_design}
    
\end{table}

\begin{table}[t!]
    \centering
    \begingroup
    \footnotesize
    \setlength{\tabcolsep}{2pt}
    \renewcommand{\arraystretch}{1.12}

    \begin{minipage}[t]{0.25\linewidth}
        \centering
        \begin{tabular*}{\linewidth}{@{\extracolsep{\fill}}lrr@{}}
            \toprule
            Init. & Epochs & gFID$\downarrow$ \\
            \midrule
            GigaTok & 50 & \textbf{4.68} \\
            Scratch & 250 & 4.82 \\
            Scratch & 350 & 4.71 \\
            \bottomrule
        \end{tabular*}
    \end{minipage}\hfill
    \begin{minipage}[t]{0.53\linewidth}
        \centering
        \setlength{\tabcolsep}{1.5pt}
        \begin{tabular*}{\linewidth}{@{\extracolsep{\fill}}lrrrrrrr@{}}
            \toprule
            Sampling & 64 & 96 & 128 & 160 & 192 & 224 & 256 \\
            \midrule
            $p=0.3$ & \textbf{6.97} & \textbf{5.68} & 5.23 & 5.10 & 4.92 & 4.99 & 5.03 \\
            $p=0.5$ & 8.91 & 6.63 & 5.72 & 5.39 & 5.11 & 5.02 & 4.74 \\
            Padding & 7.55 & 5.76 & \textbf{5.12} & \textbf{4.93} & \textbf{4.81} & \textbf{4.73} & \textbf{4.68} \\
            \bottomrule
        \end{tabular*}
    \end{minipage}\hfill
    \begin{minipage}[t]{0.19\linewidth}
        \centering
        \begin{tabular*}{\linewidth}{@{\extracolsep{\fill}}lr@{}}
            \toprule
            Teacher & gFID$\downarrow$ \\
            \midrule
            DINOv2-B & \textbf{2.66} \\
            DINOv2-L & 2.74 \\
            DINOv3-L & 2.82 \\
            \bottomrule
        \end{tabular*}
    \end{minipage}

    \par\medskip
    \begin{minipage}[t]{0.25\linewidth}
        \centering (a) Tokenizer initialization
    \end{minipage}\hfill
    \begin{minipage}[t]{0.53\linewidth}
        \centering (b) Padding vs. reweighting
    \end{minipage}\hfill
    \begin{minipage}[t]{0.19\linewidth}
        \centering (c) Teacher encoder
    \end{minipage}
    \par\medskip
    \endgroup

    \caption{Additional design choices of ReTok.
    (a) Effect of tokenizer initialization.
    (b) Comparison of padding and prefix-length reweighting. We compare redundant token padding with no padding using $p=P(k=256)\in\{0.3,0.5\}$, with the remaining lengths sampled uniformly.
    (c) Effect of the representation-alignment teacher.
    Experiments in (a,b) use ReTok-S-S with AR-B trained for 120 epochs, while (c) follows the full ReTok-S-B training protocol with AR-L trained for 300 epochs.}
    \label{tab_additional_controls}
\end{table}

\noindent \textbf{Visualize Latent Features of Tokenizer.} 
We visualize latent features from the first ViT decoder layer using PCA with 3 components. 
PCA is computed on one class and applied to reduce features to 3 dimensions for visualization.
We compare feature maps at 32 and 256 tokens in \cref{fig_heatmap} (b).
The results show that early tokens capture the global shape of the main object, while the full sequence adds detailed structure and texture.

\noindent \textbf{Design and Training Choices for ReTok.} 
We determine key design choices for ReTok. 
(1) Number of padding tokens. By default, we pad 224 redundant tokens. In \cref{tab4_design} (a), we find padding more tokens (384) or fewer tokens (64) leads to worse reconstruction and generation results. 
(2) Initial weight of hierarchical semantic regularization. As shown in \cref{tab4_design} (b), a large initial weight for 32 tokens degrades both generation and reconstruction of the tokenizer, while a small weight slightly leads to a performance drop on gFID.
(3) Step size of nested dropout. We compare two settings for sampling $k$ during nested dropout: step size 32 ($k\in\{32,64,...,256\}$) and step size 16 ($k\in\{32, 48,...256\}$). We observe in \cref{tab4_design} (c) that using a shorter step size (step size 16) performs comparable performance at full length, while degrading both reconstruction and generation at 32 tokens. 
(4) GigaTok Initialization. Training from scratch for 350 epochs, including a 50-epoch warm-up without token dropout, achieves 4.71 gFID versus 4.68 with GigaTok initialization and 50 epochs of training (\cref{tab_additional_controls} (a)), showing that pre-trained initialization is not essential. We use pre-trained GigaTok to accelerate convergence. 
(5) Redundant token padding vs. reweighting. Without padding, we set $p=P(k=256)$ to 0.3 or 0.5 and sample the remaining lengths uniformly. Increasing $p$ improves full-length generation but degrades short-prefix generation quality; redundant token padding outperforms $p=0.5$ at all reported lengths (\cref{tab_additional_controls} (b)).
(6) Alignment teacher. Under the same full ReTok-S-B and AR-L training protocol, DINOv2-L and DINOv3-L \cite{simeoni2025dinov3} do not improve upon DINOv2-B (\cref{tab_additional_controls} (c)). We therefore use DINOv2-B as the teacher encoder.

\section{Conclusion}
In this paper, we systematically analyze the generation bottleneck of current flexible tokenizers when applying naive nested dropout.
To address these issues, we present ReTok, a 1D visual tokenizer that allows flexible AR generation to achieve consistent improvement in generation as the token sequence extends.
We propose redundant token padding, hierarchical semantic regularization, and decoder fine-tuning to exploit full sequence length for better latent modeling.
We conduct extensive experiments to evaluate the effectiveness of our method.
On ImageNet 256$\times$256, our ReTok-S-B with AR-XL achieves 2.27 gFID, demonstrating its superior performance compared to other flexible and fixed-length tokenizers.
Discussions of implementation details, visual examples, and limitations are presented in the appendix.  
% We present ReTok, a 1D flexible tokenizer that mitigates the generation bottleneck of naive nested dropout.
% %
% By combining redundant token padding, hierarchical semantic regularization, and decoder fine-tuning, ReTok better exploits full token sequences.
% %
% On ImageNet 256$\times$256, ReTok-S-B with AR-XL achieves 2.27 gFID, outperforming existing tokenizers.
%%%%%%%%%%%%%%%%%%%%%%%%%%%%%%%%%%%%%%%%%%%%%%%%%%%%%%%%%%%%

\begin{ack}
This work was supported by the Rapid-Rich Object Search (ROSE) Lab, School of Electrical and Electronic Engineering, Nanyang Technological University (NTU), Singapore.
\end{ack}

\bibliographystyle{plain}
\bibliography{refs}

\clearpage
\appendix

\section{Technical appendices and supplementary material}
\begin{figure}[h]
    \centering

    % ================= 第一排：只放图片（使用 [b] 底部对齐） =================
    \begin{minipage}[t]{0.48\linewidth}
        \centering
        \includegraphics[width=\linewidth]{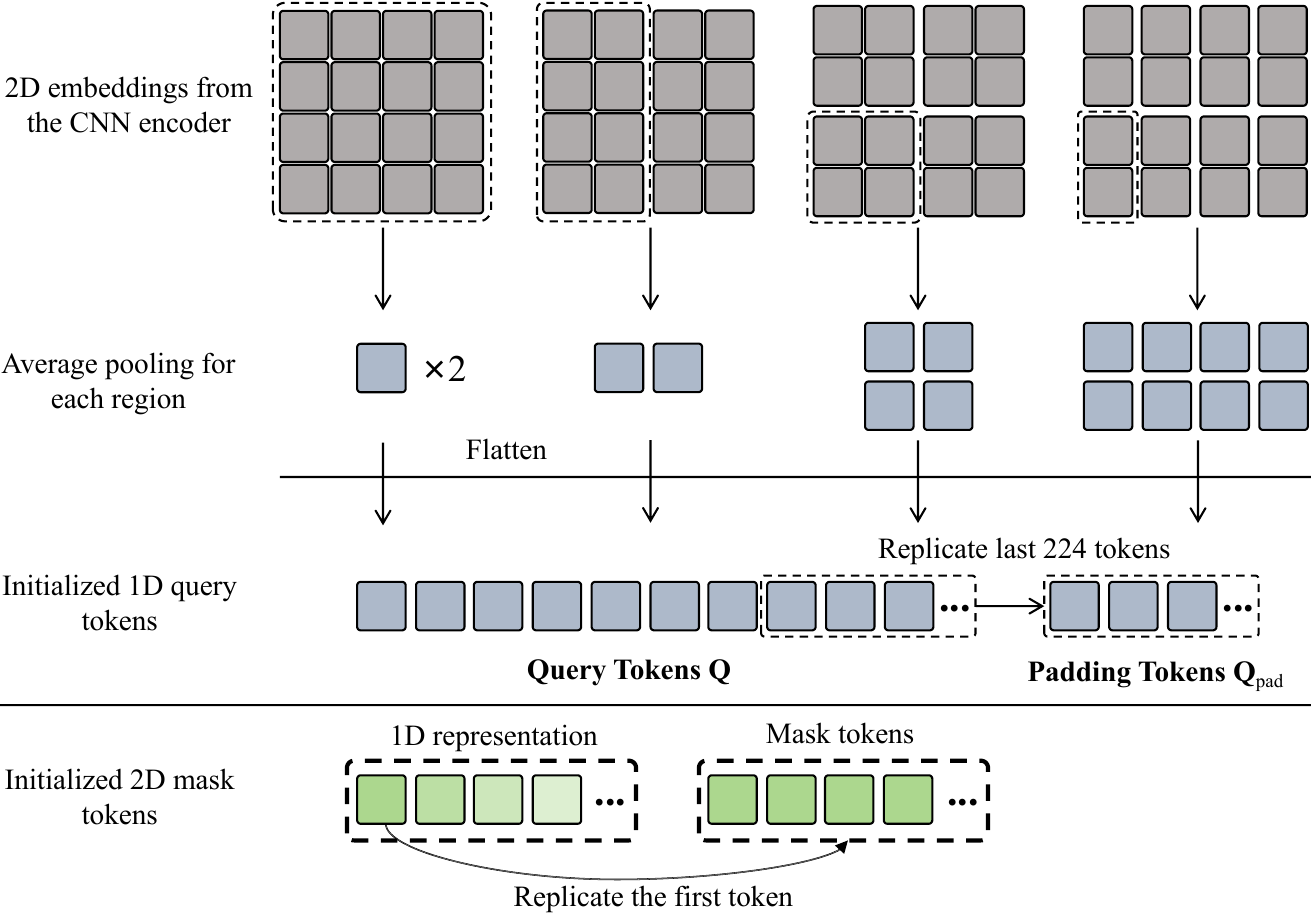}
    \end{minipage}
    \hfill 
    \begin{minipage}[t]{0.48\linewidth}
        \centering
        \includegraphics[width=0.99\linewidth]{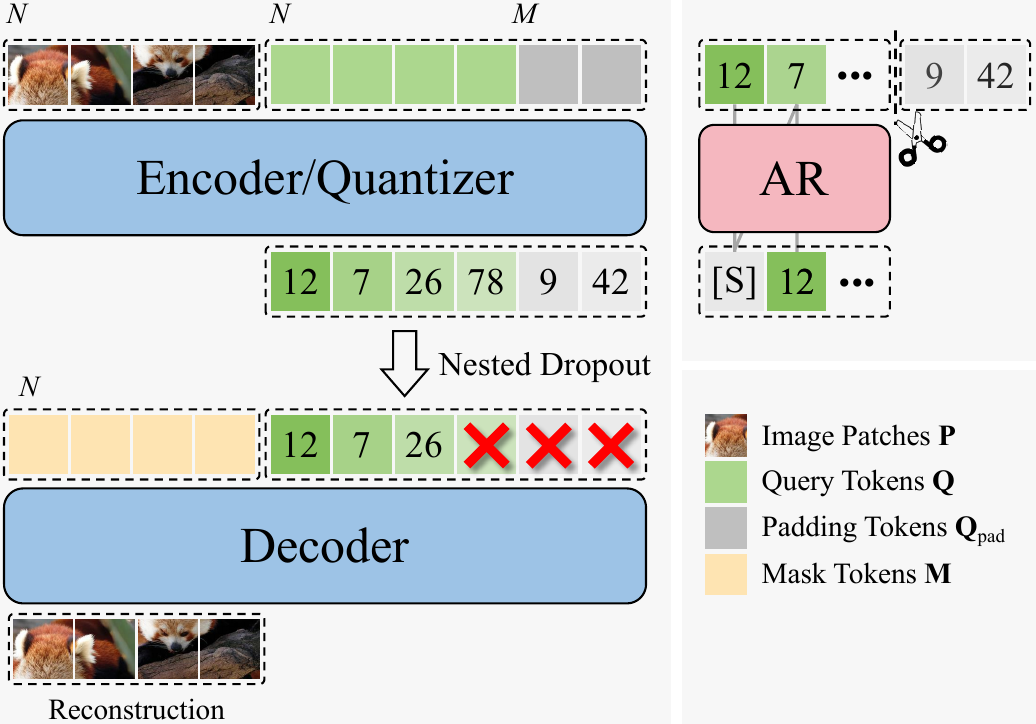}
    \end{minipage}\\ % <--- 核心：强制换行，彻底切断图片和Caption之间的垂直拉伸关联

    % ================= 第二排：只放 caption（使用 [t] 顶部对齐） =================
    \begin{minipage}[t]{0.48\linewidth}
        % 统一设置 skip 间距，确保 Caption 紧贴图片下方
        % \captionsetup{skip=4pt} 
        \caption{Illustration of initializing query tokens and mask tokens.}
        \label{fig_query}
    \end{minipage}
    \hfill
    \begin{minipage}[t]{0.48\linewidth}
        % \captionsetup{skip=4pt} 
        \caption{Illustration of ReTok.}
        \label{fig_method}
    \end{minipage}
\end{figure}
\subsection{Implementation Details of ReTok}
\label{sec_imple_detail}
We follow GigaTok \cite{xiong2025gigatok}, which adopts a hybrid architecture comprising a CNN encoder, a ViT-based Q-Former encoder \cite{li2023blip}, a ViT-based Q-Former decoder, and a CNN decoder.
First, the CNN encoder takes the $256\times 256$ image as input and downsamples it 16 times to obtain the image embedding $\mathbf{P}$ ($L=256$). 
To extract 1D representations, we employ a Q-Former with 256 query tokens $\mathbf{Q}$, which are initialized by a multi-level average pooling strategy (see \cref{fig_query}). 
For padding tokens $\mathbf{Q}_{pad}$, we replicate the last 224 query tokens, resulting in a total of 480 tokens.
These query tokens $[\mathbf{Q}, \mathbf{Q}_{pad}]$ and image embeddings $\mathbf{P}$ are fed into the ViT encoder, which consists of alternating self-attention and cross-attention blocks (where image embeddings serve as Keys/Values). We use the absolute positional embeddings for both query tokens and image embeddings.  Finally, the truncated quantized tokens $\mathbf{Z}^\prime$ are concatenated with the mask tokens $\mathbf{M}$ and processed by the ViT decoder, followed by a CNN decoder for image reconstruction. 
Notably, the mask tokens are initialized by replicating the latent representation of the first discrete token $\mathbf{Z}^\prime$. 
We present the illustration and configurations of our ReTok in \cref{fig_method} and \cref{tab_config}.
For autoregressive models, we apply LlamaGen \cite{sun2024autoregressive} with absolute positional embeddings to model the latent distribution.
To determine the optimal Classifier-Free Guidance (CFG) scale for gFID, we start from the CFG=1.0 with a step of 0.25. 

\subsection{Full Results}
We present the quantitative results of generation (gFID) in \cref{tab_full_gfid} and diverse generated samples in \cref{fig_supple_visual_examples} by using 256 tokens.
Meanwhile, progressive generation results are also presented in \cref{fig_supple_progressive_examples}. 
\begin{table}[t]
    \centering
    % 使用 \linewidth 确保表格宽度正好撑满当前行宽
    \resizebox{0.6\linewidth}{!}{
        \begin{tabular}{lcc}
        \toprule
        \textbf{Configuration} & \textbf{ReTok-S-S} & \textbf{ReTok-S-B} \\
        % 模拟图片标题下的双线/粗线效果
        \midrule[1pt] 
        
        % --- Model Section ---
        \textbf{Model} & & \\
        \hline
        Parameters & 136M & 232M \\
        Codebook Size & \multicolumn{2}{c}{16384}  \\
        Latent Dim & \multicolumn{2}{c}{8} \\
        Num. Tokens & \multicolumn{2}{c}{256} \\
        Pad. Tokens & \multicolumn{2}{c}{224} \\
        \hline
        
        % --- Training Section ---
        \textbf{Training} & & \\
        \midrule % 用实线代替虚线，分割标题与内容
        Training Epochs & 200 & 250 \\
        Batch Size & \multicolumn{2}{c}{128} \\
        Retained Sequence & \multicolumn{2}{c}{\{32,64,...,480\}} \\
        $\lambda(k)$ & 2-0.5 & 2.5-0.5 \\
        
        \midrule % 用实线代替虚线，分割 Training 下的 Optimizer
        \textbf{Training Optimizer} & & \\
        \midrule % 用实线代替虚线
        Optimizer &  \multicolumn{2}{c}{AdamW} \\ 
        Learning Rate & \multicolumn{2}{c}{1e-4} \\
        Beta & \multicolumn{2}{c}{$\beta_1=0.9, \beta_2=0.95$} \\
        Scheduler & \multicolumn{2}{c}{Cosine Decay} \\
        End Learning Rate & \multicolumn{2}{c}{1e-5} \\
        Warmup Iterations & \multicolumn{2}{c}{0} \\
        \hline
        
        % --- Fine-tuning Section ---
        \textbf{Fine-tuning} & & \\
        \midrule % 用实线代替虚线
        Fine-Tuning Epochs & \multicolumn{2}{c}{50} \\
        Batch Size & \multicolumn{2}{c}{128} \\
        Retained Sequence & \multicolumn{2}{c}{\{32,64,...,480\}} \\
        $\lambda(k)$ &\multicolumn{2}{c}{0.5} \\
        
        \midrule % 用实线代替虚线
        \textbf{Fine-tuning Optimizer} & \multicolumn{2}{c}{Same as the training} \\
        \bottomrule
        \end{tabular}
    } % resizebox 结束
    \caption{Configurations of ReTok.}
    \label{tab_config}
\end{table}
Some images generated with fewer tokens exhibit artifacts, which are mitigated as the number of tokens increases. This indicates that image complexity varies and requires different token lengths for effective representation; specifically, complex images require more tokens to achieve high-quality generation.
We also evaluate the full results for image reconstruction by using ReTok in \cref{tab_ss_recon} and \cref{tab_sb_recon}.
Since our tokenizers are trained with the token padding, we also present the image reconstruction performance using more than 256 tokens. 

\begin{table}[t]
    \centering
    \begin{tabular}{llccccccccc}
    \toprule
         Tokenizer & Generator & CFG & 256 & 224 & 192 & 160 & 128 & 96 & 64 & 32 \\
         \hline
         ReTok-S-S & LlamaGen-B & 5.75 & 4.02 & 4.02 & 4.13 & 4.34 & 4.6 &5.16 & 6.61 & 10.55 \\
         ReTok-S-S & LlamaGen-L & 2.25 & 2.92 & 3.0 & 3.23 & 3.52 & 3.8 & 4.37 & 5.47 & 8.25 \\
         ReTok-S-B & LlamaGen-L & 1.75 & 2.66 & 2.71 & 2.83 & 3.04 & 3.18 & 3.57 & 4.54 & 6.76 \\
         ReTok-S-B & LlamaGen-XL & 1.5 & 2.27 & 2.27 & 2.38 & 2.58 & 2.7 & 3.16 & 4.05 & 6.22 \\
         \bottomrule
    \end{tabular}
    \caption{Generation performance (gFID) of AR models at different tokens.}
    \label{tab_full_gfid}
\end{table}

\subsection{Additional Ablation Study}
\noindent \textbf{Solely Using Hierarchical Semantic Regularization is Not Enough.} We conduct the experiment training ReTok-S-S without the redundant token padding. 
As shown in \cref{tab_sole_sem}, relying solely on semantic regularization yields suboptimal performance and suffers from a generation bottleneck.
For example, the model achieves better generation quality with fewer tokens (e.g., 192 and 224) than with the full 256 tokens.

\begin{table}[H]
    \centering
    % 左侧表格
    \begin{minipage}[t]{0.48\linewidth}
        \centering
        \resizebox{\linewidth}{!}{
            \begin{tabular}{lcccc}
            \toprule
            Tokens & rFID$\downarrow$ & PSNR$\uparrow$ & SSIM $\uparrow$ & LPIPS$\downarrow$ \\
            \midrule
            480 & 0.79 & 21.32 & 0.702 & 0.199 \\
            448 & 0.79 & 21.32 & 0.701 & 0.199 \\
            416 & 0.79 & 21.32 & 0.702 & 0.199 \\
            384 & 0.88 & 21.06 & 0.689 & 0.208 \\
            352 & 0.96 & 20.80 & 0.678 & 0.217 \\
            320 & 0.98 & 20.71 & 0.675 & 0.218 \\
            288 & 1.01 & 20.63 & 0.670 & 0.223 \\
            256 & 1.09 & 20.32 & 0.657 & 0.232 \\
            224 & 1.16 & 19.98 & 0.643 & 0.244 \\
            192 & 1.38 & 19.51 & 0.624 & 0.260 \\
            160 & 1.67 & 19.05 & 0.604 & 0.275 \\
            128 & 1.87 & 18.72 & 0.589 & 0.292 \\
            96  & 2.39 & 18.0  & 0.559 & 0.320 \\
            64  & 3.30 & 17.21 & 0.525 & 0.353 \\
            32  & 6.10 & 15.89 & 0.466 & 0.411 \\
            \bottomrule
            \end{tabular}
        }
        \caption{Reconstruction results of ReTok-S-S at different tokens.}
        \label{tab_ss_recon}
    \end{minipage}
    \hfill 
    % 右侧表格
    \begin{minipage}[t]{0.48\linewidth}
        \centering
        \resizebox{\linewidth}{!}{
            \begin{tabular}{lcccc}
            \toprule
            Tokens & rFID$\downarrow$ & PSNR$\uparrow$ & SSIM $\uparrow$ & LPIPS$\downarrow$ \\
            \midrule
            480 & 0.79 & 21.57 & 0.707 & 0.193 \\
            448 & 0.79 & 21.57 & 0.707 & 0.193 \\
            416 & 0.8  & 21.51 & 0.706 & 0.194 \\
            384 & 0.82 & 21.37 & 0.702 & 0.197 \\
            352 & 0.85 & 21.24 & 0.697 & 0.201 \\
            320 & 0.9  & 21.06 & 0.689 & 0.207 \\
            288 & 0.94 & 20.84 & 0.68  & 0.213 \\
            256 & 1.01 & 20.57 & 0.668 & 0.222 \\
            224 & 1.08 & 20.19 & 0.652 & 0.234 \\
            192 & 1.23 & 19.72 & 0.633 & 0.249 \\
            160 & 1.42 & 19.28 & 0.614 & 0.263 \\
            128 & 1.56 & 18.89 & 0.596 & 0.279 \\
            96  & 1.92 & 18.20 & 0.567 & 0.305 \\
            64  & 2.66 & 17.43 & 0.532 & 0.340 \\
            32  & 4.72 & 16.04 & 0.469 & 0.398 \\
            \bottomrule
            \end{tabular}
        }
        \caption{Reconstruction results of ReTok-S-B at different tokens.}
        \label{tab_sb_recon}
    \end{minipage}
\end{table}

\begin{table}[H]
    \centering
    \setlength{\tabcolsep}{4pt}

    \begin{minipage}[t]{0.48\linewidth}
        \centering
        \small
        \begin{tabular}{lccc}
            \toprule
            Tokens & 256 & 224 & 192 \\
            \midrule
            With token padding & 4.68 & 4.73 & 4.81 \\
            W/o token padding & 5.08 & 4.96 & 4.93 \\
            \bottomrule
        \end{tabular}
        \caption{Ablation study on the role of hierarchical semantic regularization (S-S tokenizer). We compare ReTok with and without token padding.}
        \label{tab_sole_sem}
    \end{minipage}
    \hfill
    \begin{minipage}[t]{0.48\linewidth}
        \centering
        \small
        \begin{tabular}{lccc}
            \toprule
            Tokens & rFID$\downarrow$ & LPIPS$\downarrow$ & gFID$\downarrow$ \\
            \midrule
            32 & 9.02 & 0.415 & 14.79 \\
            16 & 17.36 & 0.507 & 24.60 \\
            \bottomrule
        \end{tabular}
        \caption{Minimum number of starting tokens (S-S tokenizer).}
        \label{tab_min}
    \end{minipage}
\end{table}
\noindent \textbf{Minimum Numbers of Starting Tokens.} 
Although our tokenizer supports a minimum of 32 tokens, we investigated starting from even fewer, such as 16 tokens.
Unfortunately, we found that 16 tokens are insufficient to effectively represent an image for both reconstruction and generation (\cref{tab_min}).
Consequently, we set 32 tokens as our starting point.

\subsection{Limitations}
\label{sec_limitations}
Our ReTok effectively addresses the generation bottleneck caused by naive nested dropout in flexible tokenizers.
However, our tokenizer mainly focuses on the 256$\times$256 generation, while its extension to higher image resolution is unclear. We leave this for future work.
Meanwhile, our tokenizer is designed for images. 
For video tokenization, flexible tokenizers are also highly desirable due to the temporal redundancy of the videos.

\begin{figure}[t!]
\centering
\setlength{\tabcolsep}{0.5pt}
\begin{tabular}{@{}cccccc@{}}
    \includegraphics[width=0.15\linewidth]{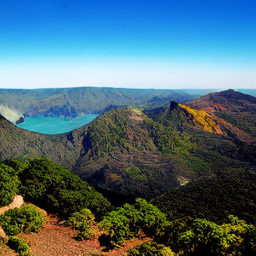} &
    \includegraphics[width=0.15\linewidth]{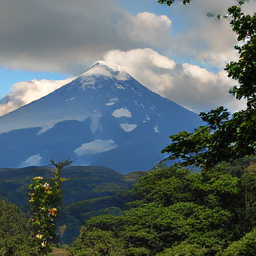} &
    \includegraphics[width=0.15\linewidth]{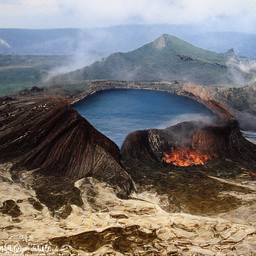} &
    \includegraphics[width=0.15\linewidth]{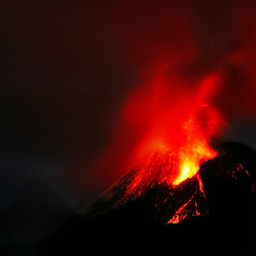} &
    \includegraphics[width=0.15\linewidth]{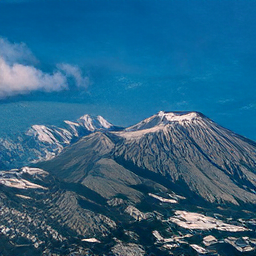} &
    \includegraphics[width=0.15\linewidth]{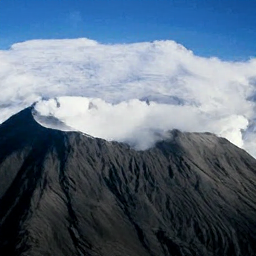}
    \\[6pt]

    \includegraphics[width=0.15\linewidth]{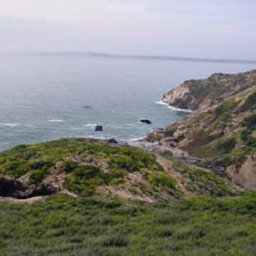} &
    \includegraphics[width=0.15\linewidth]{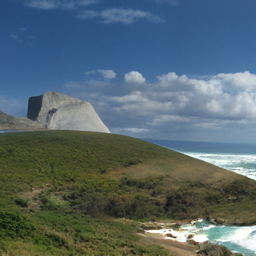} &
    \includegraphics[width=0.15\linewidth]{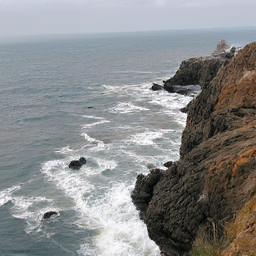} &
    \includegraphics[width=0.15\linewidth]{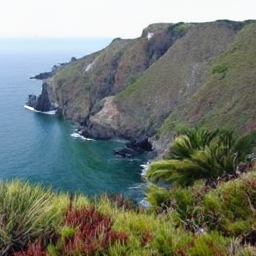} &
    \includegraphics[width=0.15\linewidth]{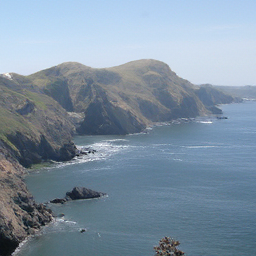} &
    \includegraphics[width=0.15\linewidth]{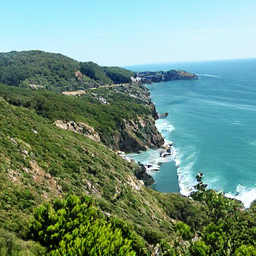}
    \\[6pt]

    \includegraphics[width=0.15\linewidth]{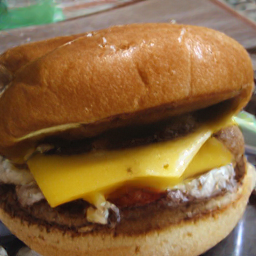} &
    \includegraphics[width=0.15\linewidth]{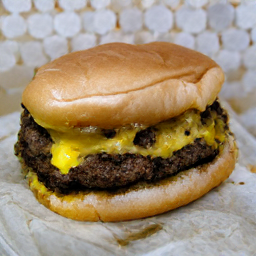} &
    \includegraphics[width=0.15\linewidth]{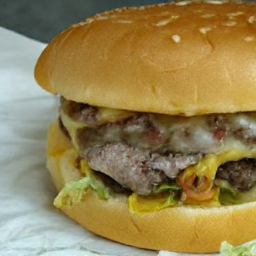} &
    \includegraphics[width=0.15\linewidth]{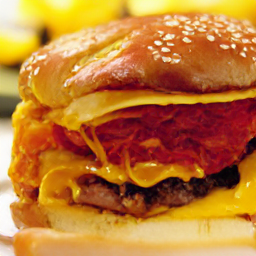} &
    \includegraphics[width=0.15\linewidth]{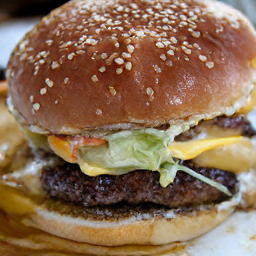} &
    \includegraphics[width=0.15\linewidth]{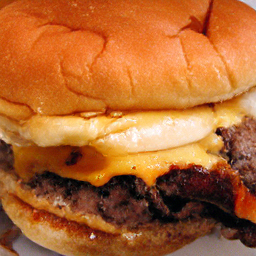}
    \\[6pt]

    \includegraphics[width=0.15\linewidth]{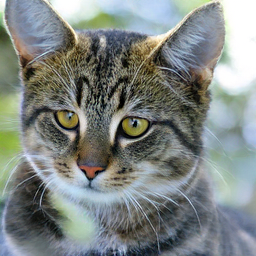} &
    \includegraphics[width=0.15\linewidth]{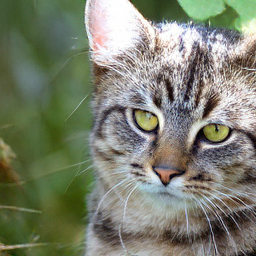} &
    \includegraphics[width=0.15\linewidth]{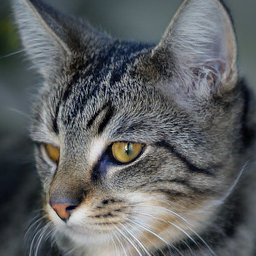} &
    \includegraphics[width=0.15\linewidth]{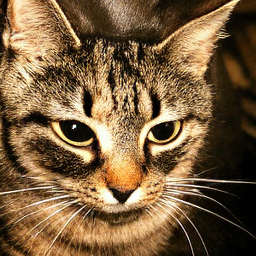} &
    \includegraphics[width=0.15\linewidth]{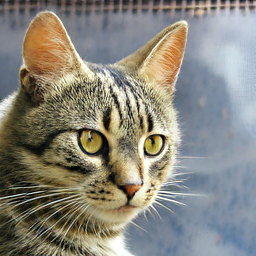} &
    \includegraphics[width=0.15\linewidth]{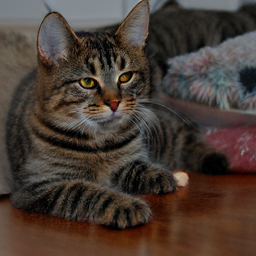}
    \\[6pt]

    \includegraphics[width=0.15\linewidth]{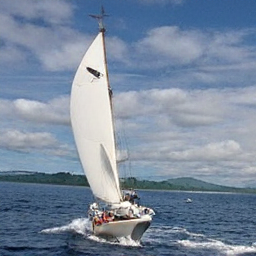} &
    \includegraphics[width=0.15\linewidth]{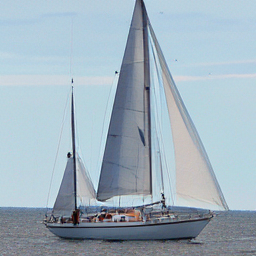} &
    \includegraphics[width=0.15\linewidth]{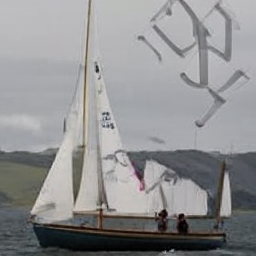} &
    \includegraphics[width=0.15\linewidth]{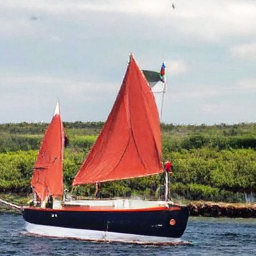} &
    \includegraphics[width=0.15\linewidth]{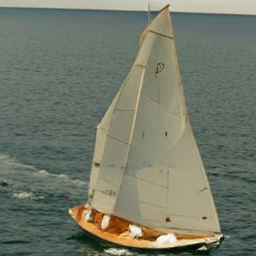} &
    \includegraphics[width=0.15\linewidth]{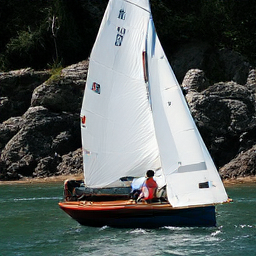}
    \\[6pt]

    \includegraphics[width=0.15\linewidth]{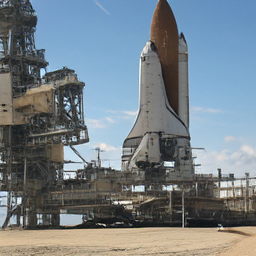} &
    \includegraphics[width=0.15\linewidth]{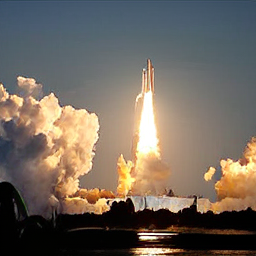} &
    \includegraphics[width=0.15\linewidth]{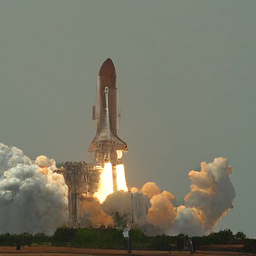} &
    \includegraphics[width=0.15\linewidth]{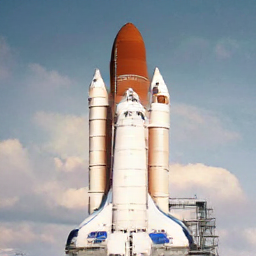} &
    \includegraphics[width=0.15\linewidth]{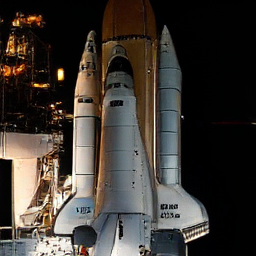} &
    \includegraphics[width=0.15\linewidth]{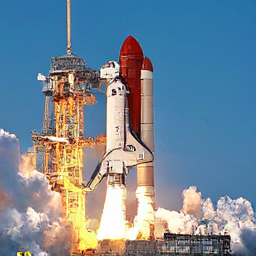}
    \\[6pt]

    \includegraphics[width=0.15\linewidth]{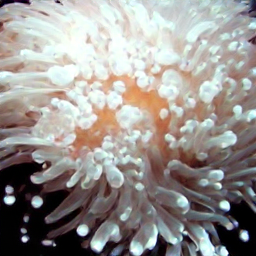} &
    \includegraphics[width=0.15\linewidth]{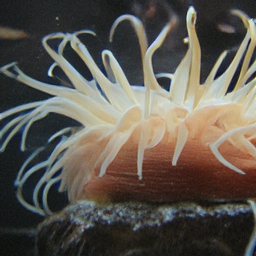} &
    \includegraphics[width=0.15\linewidth]{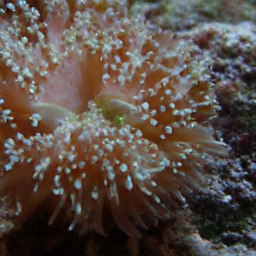} &
    \includegraphics[width=0.15\linewidth]{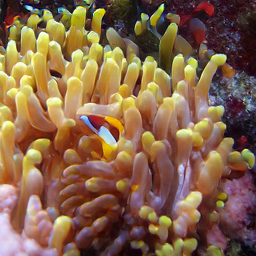} &
    \includegraphics[width=0.15\linewidth]{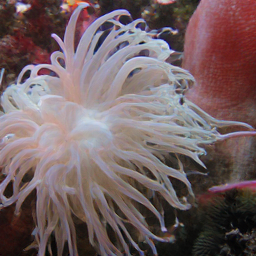} &
    \includegraphics[width=0.15\linewidth]{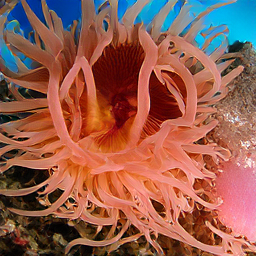}
    \\[6pt]

    \includegraphics[width=0.15\linewidth]{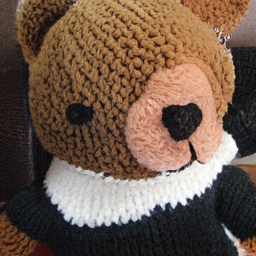} &
    \includegraphics[width=0.15\linewidth]{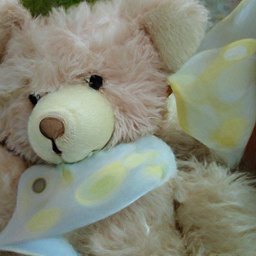} &
    \includegraphics[width=0.15\linewidth]{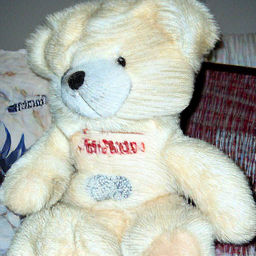} &
    \includegraphics[width=0.15\linewidth]{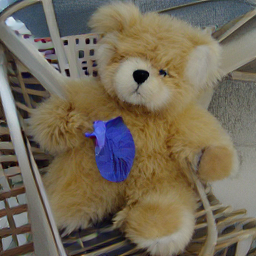} &
    \includegraphics[width=0.15\linewidth]{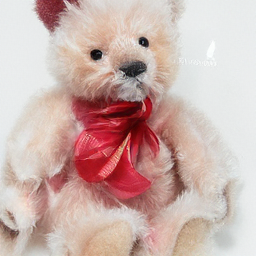} &
    \includegraphics[width=0.15\linewidth]{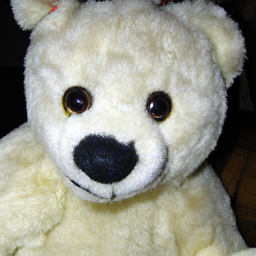}
\end{tabular}
\caption{Examples of generated images on ImageNet $256\times256$ from the ReTok-S-B + LlamaGen-XL models using 256 tokens. The classifier-free guidance is set to $4.0$.}
\label{fig_supple_visual_examples}
\end{figure}

\begin{figure}[t!]
\centering
\setlength{\tabcolsep}{0.5pt}
\begin{tabular}{@{}cccccccc@{}}
{\fontsize{6}{7}\selectfont 32 Tokens} &
{\fontsize{6}{7}\selectfont 64 Tokens} &
{\fontsize{6}{7}\selectfont 128 Tokens} &
{\fontsize{6}{7}\selectfont 256 Tokens} &
{\fontsize{6}{7}\selectfont 32 Tokens} &
{\fontsize{6}{7}\selectfont 64 Tokens} &
{\fontsize{6}{7}\selectfont 128 Tokens} &
{\fontsize{6}{7}\selectfont 256 Tokens}
\\[1pt]

\includegraphics[width=0.12\linewidth]{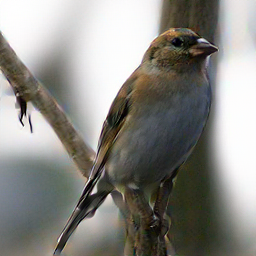} &
\includegraphics[width=0.12\linewidth]{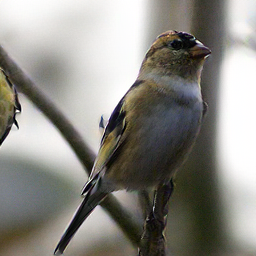} &
\includegraphics[width=0.12\linewidth]{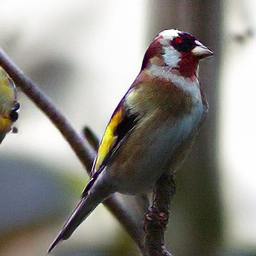} &
\includegraphics[width=0.12\linewidth]{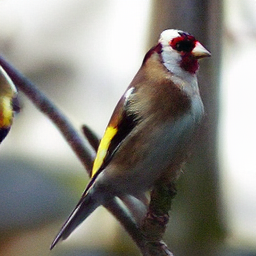} &
\includegraphics[width=0.12\linewidth]{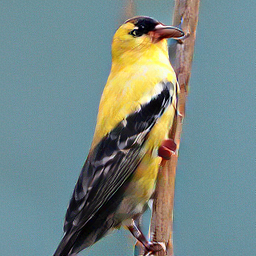} &
\includegraphics[width=0.12\linewidth]{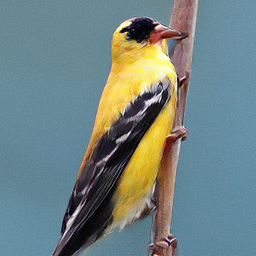} &
\includegraphics[width=0.12\linewidth]{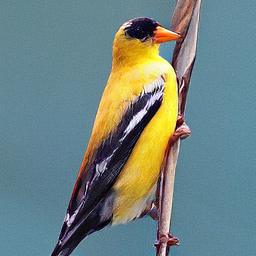} &
\includegraphics[width=0.12\linewidth]{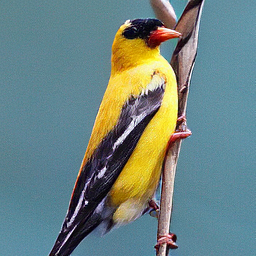}
\\[1pt]

\includegraphics[width=0.12\linewidth]{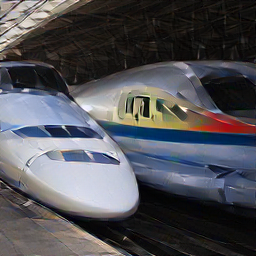} &
\includegraphics[width=0.12\linewidth]{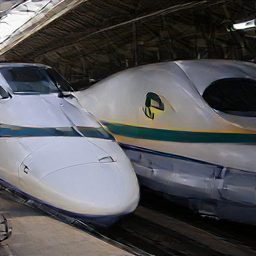} &
\includegraphics[width=0.12\linewidth]{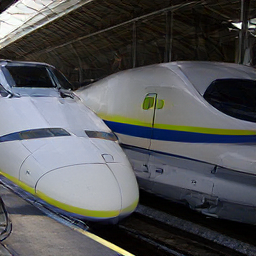} &
\includegraphics[width=0.12\linewidth]{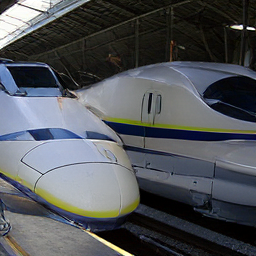} &
\includegraphics[width=0.12\linewidth]{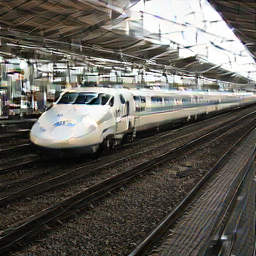} &
\includegraphics[width=0.12\linewidth]{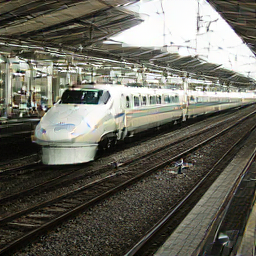} &
\includegraphics[width=0.12\linewidth]{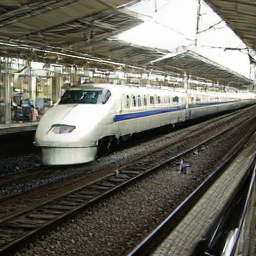} &
\includegraphics[width=0.12\linewidth]{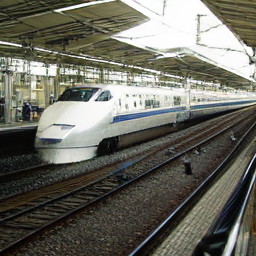}
\\[1pt]

\includegraphics[width=0.12\linewidth]{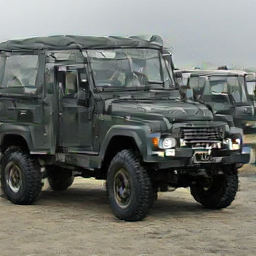} &
\includegraphics[width=0.12\linewidth]{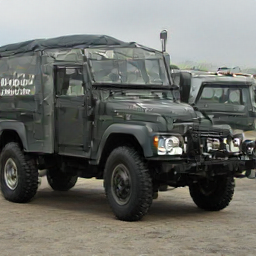} &
\includegraphics[width=0.12\linewidth]{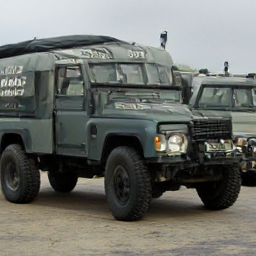} &
\includegraphics[width=0.12\linewidth]{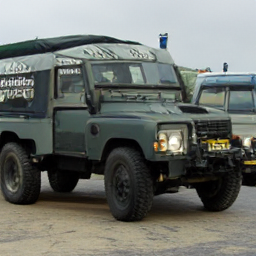} &
\includegraphics[width=0.12\linewidth]{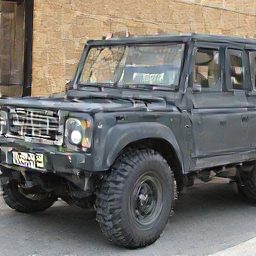} &
\includegraphics[width=0.12\linewidth]{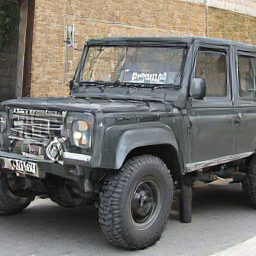} &
\includegraphics[width=0.12\linewidth]{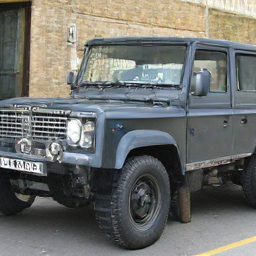} &
\includegraphics[width=0.12\linewidth]{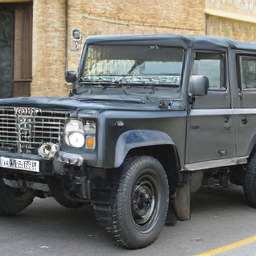}
\\[1pt]

\includegraphics[width=0.12\linewidth]{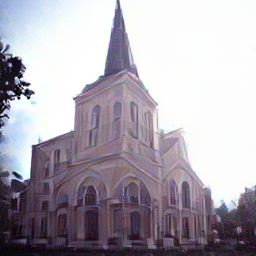} &
\includegraphics[width=0.12\linewidth]{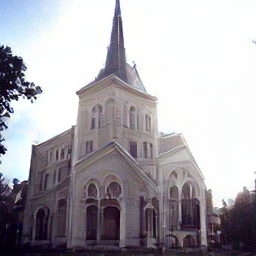} &
\includegraphics[width=0.12\linewidth]{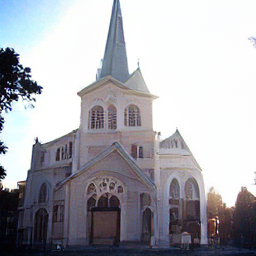} &
\includegraphics[width=0.12\linewidth]{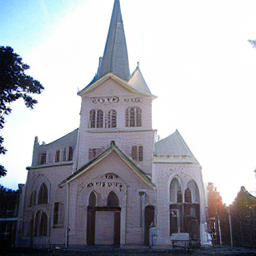} &
\includegraphics[width=0.12\linewidth]{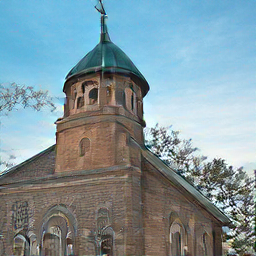} &
\includegraphics[width=0.12\linewidth]{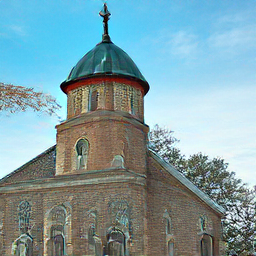} &
\includegraphics[width=0.12\linewidth]{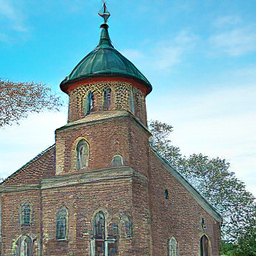} &
\includegraphics[width=0.12\linewidth]{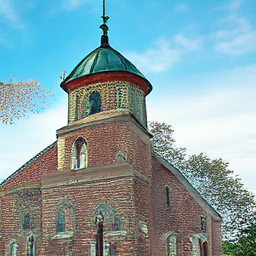}
\\[1pt]

\includegraphics[width=0.12\linewidth]{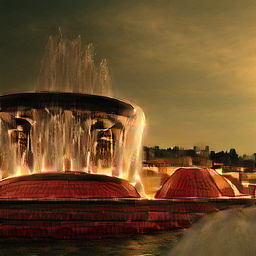} &
\includegraphics[width=0.12\linewidth]{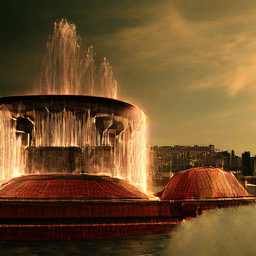} &
\includegraphics[width=0.12\linewidth]{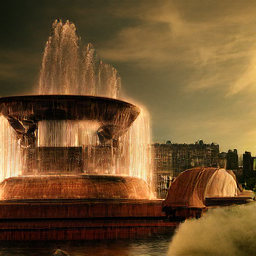} &
\includegraphics[width=0.12\linewidth]{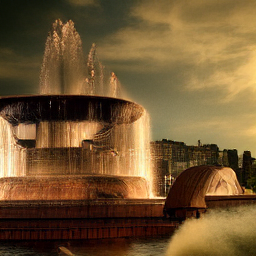} &
\includegraphics[width=0.12\linewidth]{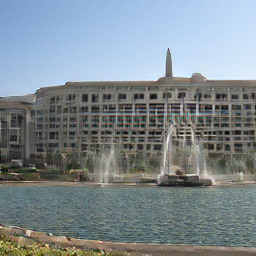} &
\includegraphics[width=0.12\linewidth]{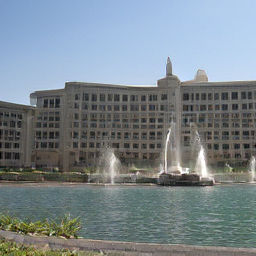} &
\includegraphics[width=0.12\linewidth]{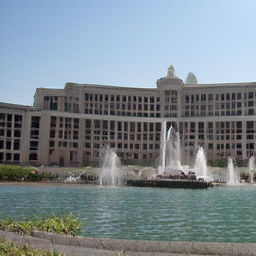} &
\includegraphics[width=0.12\linewidth]{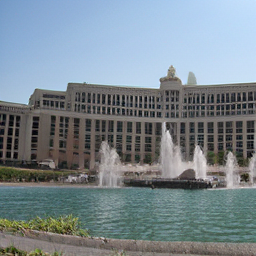}
\\[1pt]

\includegraphics[width=0.12\linewidth]{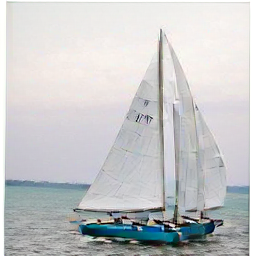} &
\includegraphics[width=0.12\linewidth]{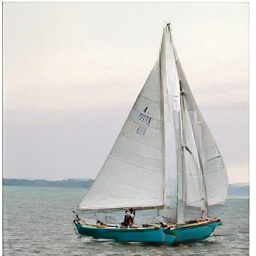} &
\includegraphics[width=0.12\linewidth]{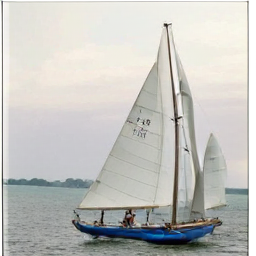} &
\includegraphics[width=0.12\linewidth]{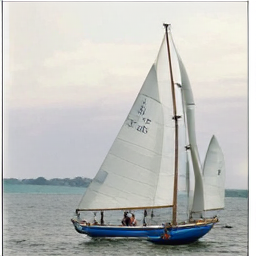} &
\includegraphics[width=0.12\linewidth]{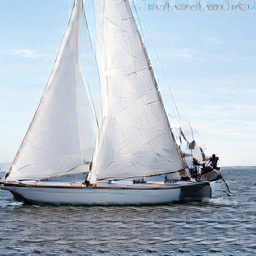} &
\includegraphics[width=0.12\linewidth]{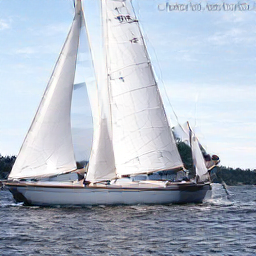} &
\includegraphics[width=0.12\linewidth]{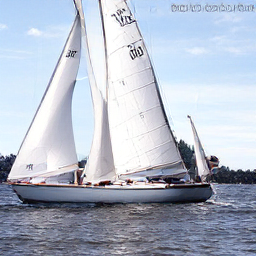} &
\includegraphics[width=0.12\linewidth]{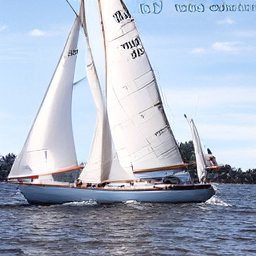}
\\[1pt]

\includegraphics[width=0.12\linewidth]{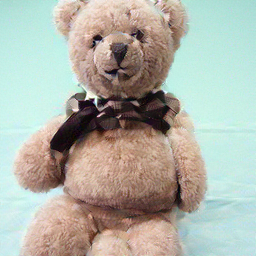} &
\includegraphics[width=0.12\linewidth]{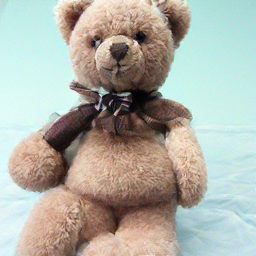} &
\includegraphics[width=0.12\linewidth]{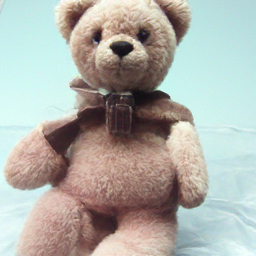} &
\includegraphics[width=0.12\linewidth]{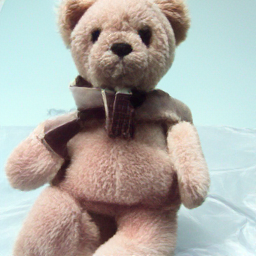} &
\includegraphics[width=0.12\linewidth]{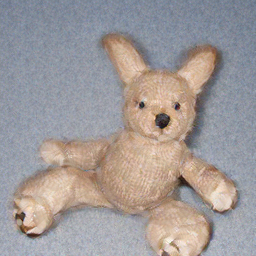} &
\includegraphics[width=0.12\linewidth]{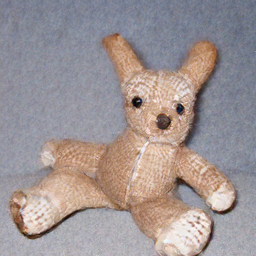} &
\includegraphics[width=0.12\linewidth]{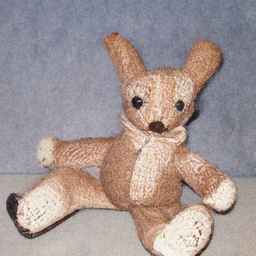} &
\includegraphics[width=0.12\linewidth]{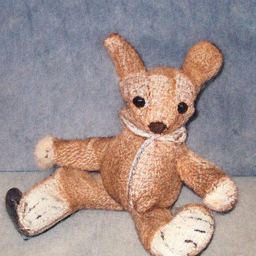}
\\[1pt]

\includegraphics[width=0.12\linewidth]{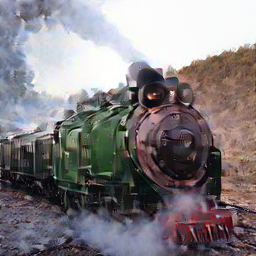} &
\includegraphics[width=0.12\linewidth]{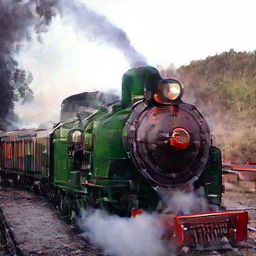} &
\includegraphics[width=0.12\linewidth]{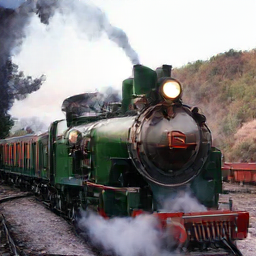} &
\includegraphics[width=0.12\linewidth]{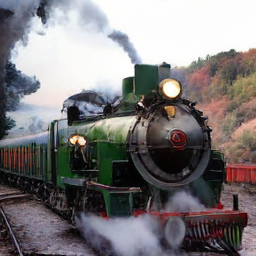} &
\includegraphics[width=0.12\linewidth]{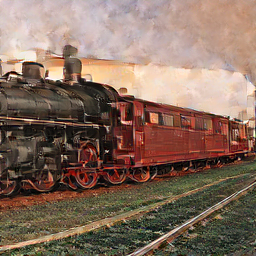} &
\includegraphics[width=0.12\linewidth]{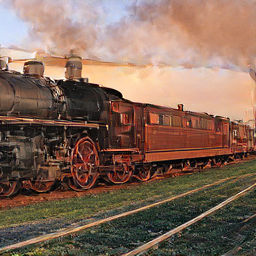} &
\includegraphics[width=0.12\linewidth]{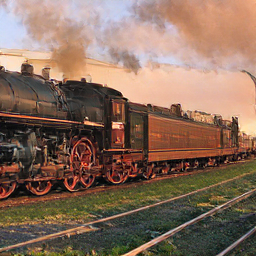} &
\includegraphics[width=0.12\linewidth]{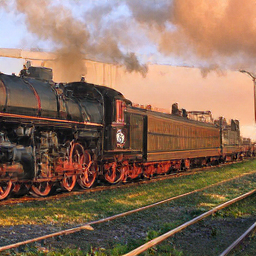}
\\[1pt]

\includegraphics[width=0.12\linewidth]{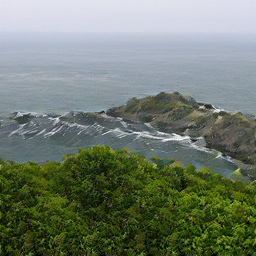} &
\includegraphics[width=0.12\linewidth]{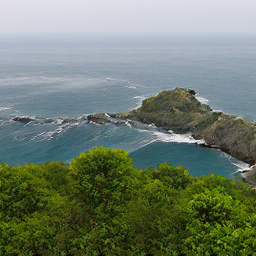} &
\includegraphics[width=0.12\linewidth]{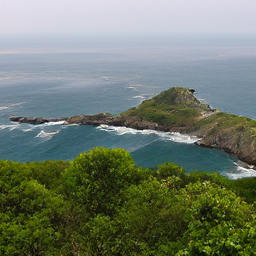} &
\includegraphics[width=0.12\linewidth]{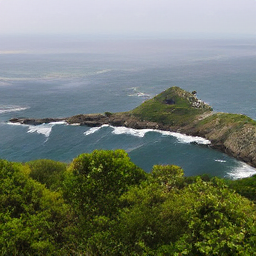} &
\includegraphics[width=0.12\linewidth]{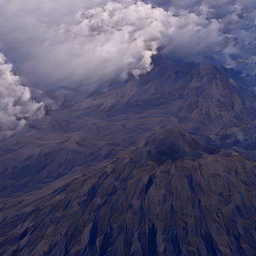} &
\includegraphics[width=0.12\linewidth]{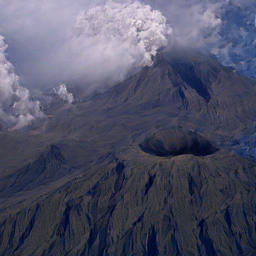} &
\includegraphics[width=0.12\linewidth]{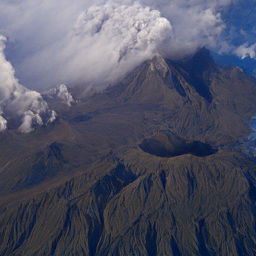} &
\includegraphics[width=0.12\linewidth]{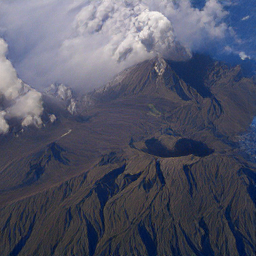}
\\[1pt]

\includegraphics[width=0.12\linewidth]{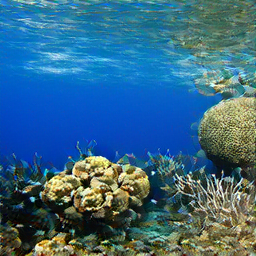} &
\includegraphics[width=0.12\linewidth]{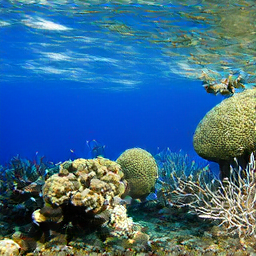} &
\includegraphics[width=0.12\linewidth]{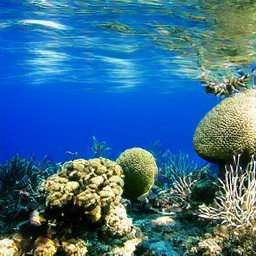} &
\includegraphics[width=0.12\linewidth]{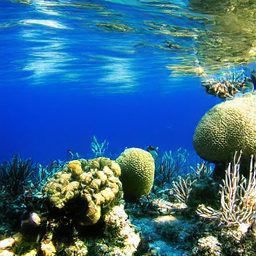} &
\includegraphics[width=0.12\linewidth]{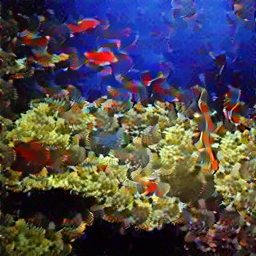} &
\includegraphics[width=0.12\linewidth]{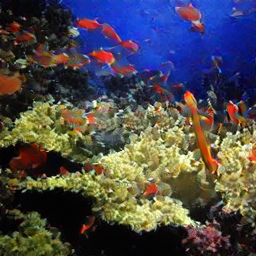} &
\includegraphics[width=0.12\linewidth]{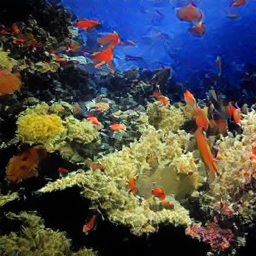} &
\includegraphics[width=0.12\linewidth]{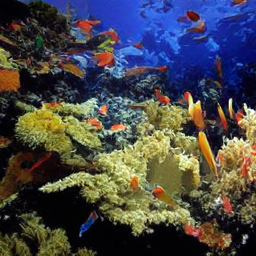}

\end{tabular}
\caption{Examples of progressive generation on ImageNet $256\times256$ from the ReTok-S-B + LlamaGen-XL models. Complex scenes require more tokens, while a small number of tokens is sufficient for simple scenes.}
\label{fig_supple_progressive_examples}
\end{figure}
%%%%%%%%%%%%%%%%%%%%%%%%%%%%%%%%%%%%%%%%%%%%%%%%%%%%%%%%%%%%

\clearpage
\newpage
\section*{NeurIPS Paper Checklist}

\begin{enumerate}

\item {\bf Claims}
    \item[] Question: Do the main claims made in the abstract and introduction accurately reflect the paper's contributions and scope?
    \item[] Answer: \answerYes{} % Replace by \answerYes{}, \answerNo{}, or \answerNA{}.
    \item[] Justification: The abstract and introduction summarize the paper's main contributions and scope. The corresponding method and experiments are provided in \cref{sec:method} and \cref{sec_exp}.

\item {\bf Limitations}
    \item[] Question: Does the paper discuss the limitations of the work performed by the authors?
    \item[] Answer: \answerYes{} % Replace by \answerYes{}, \answerNo{}, or \answerNA{}.
    \item[] Justification: The limitations are discussed in \cref{sec_limitations}

\item {\bf Theory assumptions and proofs}
    \item[] Question: For each theoretical result, does the paper provide the full set of assumptions and a complete (and correct) proof?
    \item[] Answer: \answerNA{} % Replace by \answerYes{}, \answerNo{}, or \answerNA{}.
    \item[] Justification: this paper does not present theoretical results, theorems, or formal proofs. Our contributions are methodological and empirical, focusing on flexible image tokenizer design and downstream generation evaluation.

    \item {\bf Experimental result reproducibility}
    \item[] Question: Does the paper fully disclose all the information needed to reproduce the main experimental results of the paper to the extent that it affects the main claims and/or conclusions of the paper (regardless of whether the code and data are provided or not)?
    \item[] Answer: \answerYes{} % Replace by \answerYes{}, \answerNo{}, or \answerNA{}.
    \item[] Justification: We provide the key experimental settings, including model configurations, training schedules, tokenizer settings, evaluation metrics, and datasets in \cref{sec_exp} and \cref{sec_imple_detail}. These details are sufficient to reproduce the main results supporting our claims.

\item {\bf Open access to data and code}
    \item[] Question: Does the paper provide open access to the data and code, with sufficient instructions to faithfully reproduce the main experimental results, as described in supplemental material?
    \item[] Answer: \answerNo{} % Replace by \answerYes{}, \answerNo{}, or \answerNA{}.
    \item[] Justification: We use ImageNet as the training and evaluation dataset, which is obtainable from the official source subject to its license. Code is not included in the submission, but we provide detailed implementation settings in \cref{sec_exp}.

\item {\bf Experimental setting/details}
    \item[] Question: Does the paper specify all the training and test details (e.g., data splits, hyperparameters, how they were chosen, type of optimizer) necessary to understand the results?
    \item[] Answer: \answerYes{} % Replace by \answerYes{}, \answerNo{}, or \answerNA{}.
    \item[] Justification: We provide sufficient details in the paper to understand the reported results.

\item {\bf Experiment statistical significance}
    \item[] Question: Does the paper report error bars suitably and correctly defined or other appropriate information about the statistical significance of the experiments?
    \item[] Answer: \answerNo{} % Replace by \answerYes{}, \answerNo{}, or \answerNA{}.
    \item[] Justification: We do not report error bars for the main experiments due to the high computational cost of training image tokenizers and AR models. We ensure fair comparison by using consistent training, evaluation protocols, and standard ImageNet validation metrics.

\item {\bf Experiments compute resources}
    \item[] Question: For each experiment, does the paper provide sufficient information on the computer resources (type of compute workers, memory, time of execution) needed to reproduce the experiments?
    \item[] Answer: \answerNo{} % Replace by \answerYes{}, \answerNo{}, or \answerNA{}.
    \item[] Justification: We do not provide a complete accounting of compute resources for each experiment. However, we describe the model configurations with training and evaluation settings needed to reproduce the main results.
    
\item {\bf Code of ethics}
    \item[] Question: Does the research conducted in the paper conform, in every respect, with the NeurIPS Code of Ethics \url{https://neurips.cc/public/EthicsGuidelines}?
    \item[] Answer: \answerYes{} % Replace by \answerYes{}, \answerNo{}, or \answerNA{}.
    \item[] Justification: We have reviewed the NeurIPS Code of Ethics and believe this work conforms to it. The paper focuses on standard image generation benchmarks, without involving human subjects, private data, or high-risk deployment.

\item {\bf Broader impacts}
    \item[] Question: Does the paper discuss both potential positive societal impacts and negative societal impacts of the work performed?
    \item[] Answer: \answerNo{} % Replace by \answerYes{}, \answerNo{}, or \answerNA{}.
    \item[] Justification: We do not include a broader impacts discussion. The work focuses on methodological improvements for image tokenization and generation.
    
\item {\bf Safeguards}
    \item[] Question: Does the paper describe safeguards that have been put in place for responsible release of data or models that have a high risk for misuse (e.g., pre-trained language models, image generators, or scraped datasets)?
    \item[] Answer: \answerNA{} % Replace by \answerYes{}, \answerNo{}, or \answerNA{}.
    \item[] Justification: We do not release new datasets or high-risk pretrained generative models.

\item {\bf Licenses for existing assets}
    \item[] Question: Are the creators or original owners of assets (e.g., code, data, models), used in the paper, properly credited and are the license and terms of use explicitly mentioned and properly respected?
    \item[] Answer: \answerYes{} % Replace by \answerYes{}, \answerNo{}, or \answerNA{}.
    \item[] Justification: We properly cite the datasets, models, and prior methods used in this work, and follow their corresponding licenses and terms of use.

\item {\bf New assets}
    \item[] Question: Are new assets introduced in the paper well documented and is the documentation provided alongside the assets?
    \item[] Answer: \answerNA{} % Replace by \answerYes{}, \answerNo{}, or \answerNA{}.
    \item[] Justification: This submission does not release new assets.

\item {\bf Crowdsourcing and research with human subjects}
    \item[] Question: For crowdsourcing experiments and research with human subjects, does the paper include the full text of instructions given to participants and screenshots, if applicable, as well as details about compensation (if any)? 
    \item[] Answer: \answerNA{} % Replace by \answerYes{}, \answerNo{}, or \answerNA{}.
    \item[] Justification: This paper does not involve crowdsourcing or research with human subjects.

\item {\bf Institutional review board (IRB) approvals or equivalent for research with human subjects}
    \item[] Question: Does the paper describe potential risks incurred by study participants, whether such risks were disclosed to the subjects, and whether Institutional Review Board (IRB) approvals (or an equivalent approval/review based on the requirements of your country or institution) were obtained?
    \item[] Answer: \answerNA{} % Replace by \answerYes{}, \answerNo{}, or \answerNA{}.
    \item[] Justification: This paper does not involve crowdsourcing or research with human subjects.

\item {\bf Declaration of LLM usage}
    \item[] Question: Does the paper describe the usage of LLMs if it is an important, original, or non-standard component of the core methods in this research? Note that if the LLM is used only for writing, editing, or formatting purposes and does \emph{not} impact the core methodology, scientific rigor, or originality of the research, declaration is not required.
    %this research? 
    \item[] Answer: \answerNA{} % Replace by \answerYes{}, \answerNo{}, or \answerNA{}.
    \item[] Justification: The core method development in this work does not involve LLMs as important, original, or non-standard components.

\end{enumerate}

\end{document}